\documentclass[]{article}
\usepackage{proceed2e}

\usepackage{times}
\usepackage{graphicx} % more modern
\usepackage{subfigure} 

\usepackage{hyperref}

\usepackage{qiangstyle}
\usepackage{times}

\title{Learning to Draw Samples with Amortized Stein Variational Gradient Descent}

\author{
Yihao Feng ~~~~~~~~~Dilin Wang ~~~~~~~~~Qiang Liu\\
Computer Science, Dartmouth College\\
Hanover, NH, 03755 \\
\texttt{\{yihao.feng.gr, dilin.wang.gr, qiang.liu\}@dartmouth.edu}
}

\begin{document}

\maketitle

%\footnotetext{hi}

\begin{abstract} 
We propose a simple algorithm to train stochastic neural networks
to draw samples from given target distributions for probabilistic inference. 
Our method is based on iteratively adjusting the neural network parameters 
so that the output changes along a Stein variational gradient direction \citep{liu2016stein} that maximally decreases 
the KL divergence with the target distribution. 
%to decrease the KL divergence between the output random variable and the target distribution, by mimicing a recently proposed Stein variational gradient descent. 
Our method works for any target distribution specified by their unnormalized density function, 
and can train any black-box architectures that are differentiable in terms of the parameters we want to adapt. 
%We show that our method provides efficient amortized inference approximation by applying it to two different applications. 
%One is Variational Autoencoder with black-box inference networks, and another is learning to draw samples from a set of family distributions.
We demonstrate our method with a number of applications, 
including variational autoencoder (VAE) with 
expressive encoders to model complex latent space structures, 
%complex encoder by using dropout noise, 
and hyper-parameter learning of MCMC samplers that allows Bayesian inference to adaptively improve itself 
when seeing more data.  
%We demonstrate our method using 
%Experiments consider sampling from Gaussian Mixture, Gaussian-Bernoulli Restricted Boltzmann Machine (RBM), or Bayesian classifiers posterior distributions with a black-box architecture. As a real world challenging test, we use our method to train deep generative models, and obtains realistic looking images competitive with the state-of-the-art results.
\end{abstract} 

%!TEX root = ../main.tex
\section{INTRODUCTION}

Modern machine learning increasingly relies on highly complex probabilistic models to reason about uncertainty.  
A key computational challenge is to develop efficient inference techniques to approximate, or draw samples from complex distributions. 
Currently, most inference methods, including MCMC and variational inference, are \emph{hand-designed} by researchers or domain experts. 
This makes it difficult to fully optimize the choice of different methods and their parameters, and exploit the structures in the problems of interest in an automatic way. 
The hand-designed algorithm can also be inefficient when there is a need to perform fast inference repeatedly on a large number of different distributions with similar structures. 
This happens, for example, when we need to reason about a number of observed datasets in settings like online learning or personalized prediction,  
or need fast inference as inner loops for other algorithms such as  learning latent variable models (such as variational autoencoder \citep{kingma2013auto}) or unnormalized distributions. 
Therefore, it is highly desirable to develop intelligent probabilistic inference systems that can adaptively improve their own performance to fully optimize the computational efficiency, and generalize to new tasks with similar structures. 
%
%%Specifically,
%In this work, we study 
%^A key component of such systems involve %
Developing such systems requires solving 
the following \emph{learning-to-sample} problem: 
%Denote by $p(z)$ a probability density of interest, specified up to the normalization constant, from which we want to draw samples: 
%or marginalize to estimate its normalization constant. 
\begin{problem}\label{pro:prob1}
Given a distribution with density $p(z)$ on set $\Z$ %specified up to the normalization constant, 
and a simulator $z= f(\xi;~\eta)$, such as a neural network, 
which takes a parameter $\eta$ and a random seed $\xi$ drawn from $q_0$ and outputs a value $z$ in $\Z$, 
we want to find an optimal parameter $\eta$ so that the density of the random output  $z = f(\xi;~\eta)$ with $\xi\sim q_0$ closely matches the target $p$. 
%$\eta$ and random input $\xi$, 
\end{problem}
Here, we  assume that we do not know the analytical form 
of the simulator $f(\cdot)$ (which we call inference network),  
%we assume that the simulator $f$ can be arbitrarily complex, 
and we can only query it through the output value $f(\xi;~\eta)$ and derivative $\partial_\eta f(\xi;~\eta)$ for given $\eta$ and $\xi$.  
%we do not assume that we have the access to the analytical form of $f(\eta; ~\xi)$, 
We also assume that the random seed distribution $q_0$ is unknown and we can  
only access it through the draws of the random input $\xi$; 
that is, $q_0$ can be arbitrarily complex, and can be discrete, continuous or hybrid. 
%\red{Qrevise}
% (without knowing its true distribution $q_0$)
%do not assume that the random seed distribution $q_0$ is known 
%and 
%$f(\eta; ~\xi)$ is a black-box to us, and h
%for which we only have assess to draws of the random input $\xi$ (without knowing its true distribution $q_0$),  and the output values of $f(\eta;~\xi)$ and its derivative $\partial_\eta f(\eta;~\xi)$ given $\eta$ and $\xi$.  

%Because we have no assumption on the structure of $f(\eta;~\xi)$ and the distribution of random input,
Because of the above assumption, 
we cannot directly calculate the density $q_{\eta}(z)$ of the output variable $z = f(\xi;~\eta)$; 
this makes it difficult to solve Problem~\ref{pro:prob1} using the typical variational inference (VI) methods. 
Recall that VI finds an optimal parameter $\eta$ to approximate the target $p$ with $q_{\eta}$, 
%approximates $p(x)$ using simple proposal distributions $q_\eta(x)$ indexed by parameter $\eta$, 
%and finds the optimal $\eta$ by
by minimizing the KL divergence:
\begin{align}\label{equ:kl}
\KL(q_\eta ~||~p) = \E_{q_\eta}[\log (q_\eta/p)].
\end{align}
However equation \eqref{equ:kl} requires calculating the density $q_\eta(z)$ or its derivative, which is intractable by our assumption (and is called implicit models in \citet{mohamed2016learning}). 
This holds true even when Monte Carlo gradient estimates \citep{hoffman2013stochastic} and the reparametrization trick \citep{kingma2013auto} are applied. 
This requirement of calculating $q_\eta(z)$ %causes a hurdle  
makes it difficult for practitioners to use variational inference in an automatic way,  
especially in domains where it is critical to use 
expressive inference networks to achieve good  
approximation quality. 
%are interested in using expressive 
%inference network, but does not 
%$q_\eta$ but does not  
%In fact, it is this requirement of calculating $q_\eta(x)$ that has been the major constraint for the designing of state-of-the-art variational inference methods with rich approximation families; 
%the recent successful algorithms \citep[e.g.,][to name only a few]{rezende2015variational,tran2015variational,ranganath2015hierarchical}  have to handcraft special variational families 
%to ensure the computational tractability of $q_\eta(x)$ and simultaneously obtain high approximation accuracy, 
%which require substantial mathematical insights and research effects. 
Methods that do not require to explicitly calculate $q_\eta(z)$, 
referred to as \emph{wild variational inference}, or \emph{variational programming} \citep{operator},  
can significantly simplify the design and expand the applications of VI methods, 
allowing practioners to focus more on choosing proposals that work best with their specific tasks. %, making the design and applications of VI much easier. 
%}
%We should distinguish 
%\blue{
%We use the term \emph{wild variational inference} to refer to variants of variational methods that require no tractability on density $q_\eta(z)$, 
%which is referred as variational programming in \citet{operator}. %\emph{wild variational inference}, 
%to distinguish with the
\begin{comment}
this is orthogonal to 
 \emph{black-box variational inference} \citep{ranganath2013black}
which refers to methods that work for generic target distributions $p(z)$ without gradient calculation, or significant model-by-model consideration (but still require to calculate the proposal density $q_\eta(z)$). 
\end{comment}
%\citet{operator inference paper}  uses \emph{variational programing} to refer a similar setting. 
%In \citet{tran2017deep}, distributions $q_\eta$ defined by simulator $f(\xi; ~\eta)$ is called implicit models.  
%}

\begin{figure}[t]
   \centering
   \scalebox{1}{
   \includegraphics[width=.45\textwidth]{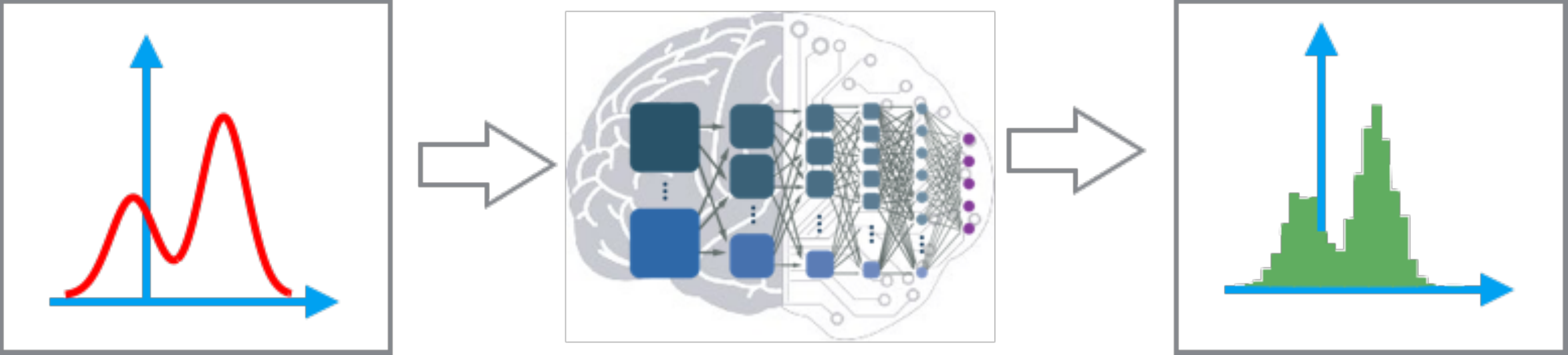}%neuralbayesian} % requires the graphicx package
   }
   \begin{picture}(0,0)(0,0)
   \put(-230,-10){\figcapsize \emph{Given distribution}}
      \put(-145,-10){{\figcapsize \emph{Inference network}}}
%      \put(-180,-10){{\figcapsize \emph{by amortized SVGD}}}      
%   \put(-180,0){\begin{tabular}{c} {\figcapsize Black-box sampler} \\[-1\baselineskip] {\figcapsize trained by amortized SVGD}    \end{tabular}}
   \put(-45,-10){\figcapsize \emph{Samples}}
   \end{picture}\\[.5em]
   \caption{
   \figcapsize   
   Wild variational inference allows us 
   to train general stochastic neural inference networks to learn to draw (approximate) samples from the target distributions, 
%   to approximate target distributions with arbitrary stochastic neural inference networks 
%   that may have computationally intractable density functions.
without restriction on the computational tractability of the density function of the neural inference networks.   
   }
%   Our methods ``learn to draw samples'',  
%   constructing black-box neural samplers for given distributions. 
%   It allows us to automatize the hyper-parameter tuning of Bayesian inference, 
% speed up the inference inner loops of learning algorithms, 
%and eventually replace hand-designed inference algorithms with more efficiently one that is trained on past tasks and is improved adaptively over time.}
% Our new amortized SVGD algorithm allows us to perform ``meta-Bayesian inference'', constructing functions that learns to produce posterior samples based on prior and data, significantly improving the efficiency of Bayesian inference at large scale.}
%   Our new amortized SVGD algorithm allows us to perform ``meta-Bayesian inference'', constructing functions that learns to produce posterior samples based on prior and data, significantly improving the efficiency of Bayesian inference at large scale.}
   \label{fig:example}
\end{figure}
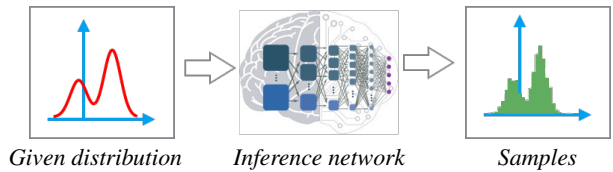

A similar problem also appears in importance sampling (IS), 
where it requires calculating the IS proposal density $q(z)$ in order to calculate the importance weight $w(z) = p(z)/q(z)$. 
However, %it is been shown that it is possible to develop %\emph{black-box importance sampling}  
there exist methods that use no explicit information of the proposal $q(z)$, and, seemingly counter-intuitively, give better asymptotic variance or converge rate than the typical IS that uses the proposal information \citep[e.g.,][]{liu2016black, briol2015probabilistic, henmi2007importance, delyon2014integral}.
% at least in the asymptotic sense. 
%This phoemnomin was first observed 
Discussions on this phenomenon date back to \citet{o1987monte}, who 
argued that ``Monte Carlo (that uses the proposal information) is fundamentally unsound'', % for violating the Likelihood Principle, 
and developed Bayesian Monte Carlo \citep{o1991bayes} as 
an instance that uses no information of proposal $q(z)$, yet gives better convergence rate than the typical Monte Carlo $\Od(n^{-1/2})$ 
convergence rate 
\citep{briol2015probabilistic}. 
Despite the substantial difference between IS and VI, 
these results intuitively suggest the possibility of developing efficient variational inference without using the information of density function $q(z)$ explicitly. 

\paragraph{Main Idea} 
In this work, we study a simple algorithm for Problem~\ref{pro:prob1}, 
motivated by a recent Stein variational gradient descent (SVGD) algorithm. 
Briefly speaking, 
SVGD is a nonparametric functional gradient descent algorithm which solves $\min_{q} \KL(q ~|| ~p)$ without parametric assumption on $q$,
and approximates the functional gradient, called the Stein variational gradient, using a set of samples (or particles) $\{z_i\}_{i=1}^n$ which iteratively evolves.  
See Section~\ref{sec:svgd}
for more introduction on SVGD. 
We can then view Problem~\ref{pro:prob1}
as a constrained optimization problem, % that minimizes 
\begin{align}
\min_q \KL(q ~|| ~p), ~~~~ s.t. ~~~ q = q_\eta ~~\text{for some $\eta$},   
\end{align}
which motivates us to develop a project gradient like algorithm that 
iteratively calculates the Stein variational gradient, and projects it to the finite dimensional $\eta$-space to update parameter $\eta$. 
At the convergence, the samples drawn from $q_\eta$ 
reach the equilibrium state of SVGD, and hence form 
a good approximation of $p$. 
We can view this method as ``amortizing'' or distilling the SVGD dynamics 
using parametric family $q_\eta$ or its inference network $f(\xi; ~\eta)$
and call it amortized SVGD. 
See %Section~\ref{sec:svgd} and 
Section \ref{sec:amortizedsvgd} for the detailed description of amortized SVGD.  
%We introduce the background of SVGD in Section~\ref{} 
%and amortze

Our algorithm provides a simple approach for 
the wild inference problem in Problem 1, enabling wide application in approximate learning and inference. 
We explore two examples in this paper. 
In Section~\ref{sec:vae}, 
we apply amortized SVGD to learn complex encoder functions in variational autoencoder (VAE), 
allowing it to 
model complex latent variable space.  
In Section~\ref{sec:trainlangevin}
we use amortized SVGD to learn hyper-parameters of MCMC samplers, 
which allows us to adaptively improve the efficiency of Bayesian computation 
when performing a large number of similar tasks.   

\paragraph {Related Work} 

%There is been a raising interesting in 
There has been a number of very recent work 
that studies advanced variational inference methods that do not require explicitly calculating $q_\eta(\z)$ (Problem 1). 
This includes 
adversarial variational Bayesian \citep{mescheder2017adversarial} which approximates 
the KL divergence with a density ratio estimator, operator variational inference \citep{operator} which replaces
the KL divergence with an alternative operator variational objective 
that is equivalent to Stein discrepancy (see 
Appendix~\ref{sec:ksd} for more discussion), 
and amortized MCMC \citep{li2017approximate} which
proposes to amortize arbitrary 
MCMC dynamics to make it applicable to discrete models and gradient-free settings.

%for which it is necessary to use a population-based two-sample discrepacncy, minimized adverse, to perform the projection step. 
%and use a population  
%that are applicable to discrete models and gradient-free settings. 
The key advantage of our method is its simplicity. 
%All the above algorithms 
both \citet{mescheder2017adversarial} and \citet{operator} 
require training some type of auxiliary networks, 
while our main algorithm (Algorithm~\ref{alg:alg1} with update \eqref{equ:update11})  
is very simple, and is essentially a generalization of the typical gradient descent rule that replaces the typical gradient with the stein variational gradient. 
%One well exception is a very recent work \citep{operator} that also avoids calculating $q_\eta(\z)$ and hence works for general inference networks; their method is based on a similar idea related to Stein discrepancy \citep{liu2016kernelized, oates2014control, chwialkowski2016kernel, gorham2015measuring}, for which we provide a more detailed discussion in Section \ref{sec:amortizedksd}. 
%\end{comment}
%Recent works\citep{operator,li2017approximate,tran2017deep,mescheder2017adversarial} that also avoid calculating $q_\eta(\z)$ and hence work for general inference networks.%\red{More discussion}

It is also possible to adopt the auxiliary variational inference methods \citep[e.g.,][]{agakov2004auxiliary,salimans2015markov} 
to solve Problem 1 by treating $\xi$ as a hidden variable. 
however, in order to frame a tractable joint distribution $p(z, \xi)$, one would need to add additional noise on $z$, e.g., assume $z = f(\xi; ~\eta) +  \normal(0, \sigma^2)$ with a careful choice of noise variance $\sigma$, and also need to introduce an additional auxiliary network 
to approximate $q(\xi ~|~ z)$, which makes the algorithm more complex than ours. % the auxiliary variables. %that introduces computational cost. 
% thatintroduce in
%where the noise variance $\sigma$ should be chosen small, but would cast stablizity problme. 
%where
%the variational distribution $q_\eta(\z)$ can be represented as a hidden variable model.  
%In particular, \cite{salimans2015markov} used the auxiliary variational approach to 
%%leverage MCMC as a variational approximation. , 
%These approaches, however, still require to write down the likelihood function on the augmented spaces, and need to introduce an additional inference network related to the auxiliary variables. 
%\red{
% }
% understand how

The idea of amortized inference \citep{gershman2014amortized}
 has been recently applied in various domains of probabilistic reasoning, 
 including both amortized variational inference \citep[e.g.,][]{kingma2013auto, jimenez2015variational} 
and date-driven designs of Monte Carlo based methods \citep[e.g.,][]{paige2016inference}, 
 to name only a few. 
 %\citet{snelson2005compact, balan2015bayesian}
 \citet{balan2015bayesian} 
  also explored the idea of amortizing or distil MCMC samplers to obtain compact fast posterior representation. 
Most of these methods uses typical variational inference methods and hence need to use simple $q_\eta$ to ensure tractability. %explicitly calculate $q_\eta(\z)$ (or its gradient). 
  
There is a large literature on traditional adaptive MCMC methods \citep[e.g.,][]{andrieu2008tutorial, roberts2009examples} which can be used to adaptively adjust the proposal distribution of MCMC by exploiting the special theoretical properties of MCMC (e.g., by minimizing the autocorrelation). 
Our method is simpler, more generic, and works efficiently in practice thanks to the use of gradient-based back-propagation. 
%Recently, 
{Finally, 
connections between stochastic gradient descent and variational inference have been discussed and exploited in 
\citet{mandt2016variational, maclaurin2015early}. }
\section{STEIN VARIATIONAL GRADIENT DESCENT}
\label{sec:svgd}
%Stein variational gradient descent (SVGD) \citep{liu2016stein} 
%is a general purpose Bayesian inference algorithm motivated by 
%optimal transport \citep[e.g.,][]{villani2008optimal}, 
%Stein's method \citep{stein1972, barbour2005introduction} and kernelized Stein discrepancy \citep{liu2016kernelized, chwialkowski2016kernel, oates2014control}. 
%It uses an efficient \emph{deterministic} gradient-based update to iteratively evolve a set of particles $\{x_i\}_{i=1}^n$ to minimize the KL divergence with the target distribution. 
%SVGD has a simple form that reduces to the typical gradient ascent for maximizing $\log p$ when using only one particle $(n=1)$, and hence can be easily combined with the successful tricks for gradient optimization, including stochastic gradient, adaptive learning rates (such as adagrad), and momentum. % making large scale Bayesian inference both much easier and more efficient. 
Stein variational gradient descent (SVGD) \citep{liu2016stein} is a 
nonparametric variational inference algorithm that 
iteratively transports a set of particles $\{z_i\}_{i=1}^n$ 
to approximate the target distribution $p$ by performing a type of functional gradient descent on the KL divergence.  
We give a quick overview of the SVGD in this section.  
% \emph{deterministic} gradient-based update to iteratively evolve a set of particles $\{x_i\}_{i=1}^n$ to

Let $p(z)$ be a positive density function on $\R^d$ which we want to approximate with a set of particles 
$\{ z_i\}_{i=1}^n$. 
SVGD starts with a set of initial particles $\{ z_i\}_{i=1}^n$, 
and 
% initializes the particles by sampling from some simple distribution $q_0$, and 
updates the particles iteratively by 
%performs iterative updates of form 
\begin{align}\label{equ:xxii}
z_i  \gets z_i +  \epsilon \ff(z_i),  ~~~~ \forall i = 1, \ldots, n,  
\end{align}
where $\epsilon$ is a step size, and 
$\ff\colon \RR^d \to \RR^d$ is a velocity field which should be chosen to push the particle distribution closer to the target distribution. 
Assume the current particles are drawn from a distribution $q$, and 
%Let $q$ be the density of the current particles,
let $q_{[\epsilon\ff]}$ be the distribution of the updated particles $z^\prime = z + \epsilon \ff(z)$ when $z \sim q$. 
The optimal choice of $\ff$ can be framed as the following optimization problem:
%$$
%\ff =  \argmax_{\ff \in \F}  \bigg\{\KL(q ~||~ p )   - \KL(q_{[\epsilon \ff]} ~||~ p),\bigg\}
% - \frac{d}{d\epsilon} \KL(q_{[\epsilon\ff]} ~|| ~ p) \big |_{\epsilon = 0}  \bigg\}, 
%$$
%that is, 
%Assuming the step size $\epsilon$ is small,  we can approximate the %
%decreasing rate of KL divergence with its gradient w.r.t. $\epsilon$ under $\epsilon = 0$, 
%``particle gradient direction'' %  perturbation direction, or velocity field, 
%which roughly speak, is the gradient of $\KL(q_\vx~||~ p)$ w.r.t. $\vx$, 
%chosen to maximumly decrease the KL divergence between the distribution of particles and the target distribution, in the sense that  
%decreases with the fastest speed in the sense that 
\begin{align}\label{equ:ff00}
\ff^* =   \argmax_{\ff \in \F}  \bigg\{  -   \frac{d}{d\epsilon} \KL(q_{[\epsilon\ff]} ~|| ~ p)~ \bigg |_{\epsilon = 0}  \bigg\}, 
%\frac{1}{\epsilon}\{ \KL(q_{t+1}~||~ p )  -  \KL(q_{t}~||~ p) \}, 
\end{align}
that is, $\ff$ should yield a maximum decreasing rate on the KL divergence between the particle distribution and the target distribution. 
Here, $\F$ is 
a function set that includes 
the possible velocity fields and is chosen to be  
%where $q_{[\epsilon \ff]}$ denotes the density of the updated particle $x^\prime = x + \epsilon \ff(x) $ when the density of the original particle $x$ is $q$, and $\F$ is the set of perturbation directions that we optimize over. 
%We choose $\F$ to be
the unit ball of a vector-valued reproducing kernel Hilbert space (RKHS) $\H = \H_0 \times \cdots \times \H_0$,
where  $\H_0$ is a RKHS formed by scalar-valued functions associated with a positive definite kernel $k(z,z')$, that is, 
$\F = \{\ff \in \H \colon ||\ff||_\H\leq  1 \}$. 
This choice of $\F$ allows us to 
consider velocity fields in infinite dimensional function spaces while still obtaining a closed form solution. 
\citet{liu2016stein} showed that 
%Critically, 
%the gradient of the KL divergence 
the objective function 
in \eqref{equ:ff00} equals a simple linear functional of $\ff$: 
%allowing us to obtain a closed form solution for the optimal $\ff$.  \citet{liu2016stein} showed that 
%A key observation is that the objective in \eqref{equ:ff00} is a linear functional of $\ff$, in fact, we have
\begin{align}\label{equ:klstein00}
&~ - \frac{d}{d\epsilon} \KL(q_{[\epsilon\ff]} ~|| ~ p) \big |_{\epsilon = 0}  = \E_{x\sim q}[\sumstein_p \ff(x)], \\[.5\baselineskip]
&~~\text{with}~~~ \sumstein_p \ff(z)  = \nabla_x \log p(z) ^\top \ff (z)+ \nabla_z^\top \ff(z),  
\end{align}
where $\sumstein_p$ is a linear operator acting on a velocity field $\ff$ and returns a scalar-valued function; % $\stein_p \ff(x)$. 
$\sumstein_p$ is called the Stein operator in connection with Stein's identity which shows that 
the RHS of \eqref{equ:klstein00} equals zero if $p = q$: 
\begin{align}\label{equ:steinid}
\E_{p}[\sumstein_p \ff] =\E_{p}[ \nabla_z \log p ^\top \ff + \nabla_z \cdot \ff] = 0. 
\end{align}
This is a result of integration by parts assuming the values of $p(z)\ff(z)$ vanish on the boundary of the integration domain.   
%Obviously, $\F$ should be taken as broad as possible, best with infinite dimension, while still allows tractable solution. 
%
%Critically, we show that derivative in \eqref{equ:f00} yields a closed form representation:
%$$  \frac{d}{d\epsilon} \KL(q_{[\epsilon\ff]} ~|| ~ p) \big |_{\epsilon = 0}  = \E_{x\sim q} [\trace(\stein_p \ff(x))], $$Critically, we show that $\F$ to be the unit ball of a reproducing kernel Hilbert space (RKHS) $\H$
%
%
%%$\KL(q_{\vx^t} ~||~ p)$ between the distribution between the distribution $q_{\vx}$ of particles $\vx^t$ and the target distribution decreases with the fastest speed. 
%We take $\F$ to be the unit ball of a reproducing kernel Hilbert space (RKHS) $\H$ associated with positive definite kernel $k(x,x')$. 
%Thanks to an important connection with Stein's identity and kernelized Stein discrepancy, we show that the optimal $\phi$ is given by 
%$\ff(x)$ in RKHS $\H$ associated with positive definite kernel $k(x,x')$, in which case the optimal choice of $\ff(\vx)$ is 
Therefore, the optimization in \eqref{equ:ff00} reduces to 
\begin{align}\label{equ:ksd}
\!\!\!\!\S(q || p) \overset{def}{=} \max_{\ff \in \H} \{ \E_{z\sim q} [\sumstein_p \ff(z)]  ~~~s.t.~~~~ ||\ff ||_{\H} \leq 1\}, 
\end{align}
where $\S(q ~||~ p)$ is the kernelized Stein discrepancy defined in \citep{liu2016kernelized, chwialkowski2016kernel}, which equals zero if and only if $p = q$ under proper conditions
that ensure the function space $\H$ is rich enough.  

Observe that \eqref{equ:ksd} 
is ``simple'' in that it is a linear functional optimization on a 
unit ball of a Hilbert space. 
Therefore, it is not surprise to derive a closed form solution:  
%which should have 
%Importantly, 
%In fact, \citet{liu2016stein} showed that it has a closed form solution: 
%the optimal solution of \eqref{equ:ksd} yields a closed form% for the optimal solution: of \eqref{equ:ff00}:
 %Further,  it turns out that optimizing \eqref{equ:klstein00} in the unit ball of $\H$ coincides with the variational form of the kernelized Stein discrepancy in \citet{liu2016kernelized},  giving a closed form for the optimal solution of \eqref{equ:ff00}: 
\begin{align}\label{equ:ffss} 
\ff^*(\cdot) \propto  \E_{z\sim q}[\nabla_z \log p(z)k(z,\cdot) + \nabla_x k(z,\cdot)], 
\end{align} 
where $k(z,z')$ is the positive definite kernel associated with RKHS $\H_0$. 
See \citet{liu2016kernelized} for the derivation. %\citep{liu2016kernelized, chwialkowski2016kernel}
We call $\ff^*$ the Stein variational gradient direction since it provides the  
optimal direction for pushing the particles towards the target distribution $p$.

%This $\ff^*$ can be treated as a gradient of the KL divergence. 
In the practical SVGD algorithm, 
we start with a set of initial particles, 
calculate its corresponding $\ff^*$ by replacing the expectation under $q$ with the empirical average of particles, 
and use it to update the particles: 
%that is, 
%the current particles $
%\{x_i\}_{i=1}^n$,  SVGD admits a simple form of update: 
\begin{align}\label{equ:update11}
%\begin{split}
%x_i ~ \gets ~ x_i  ~  + ~ \epsilon  \hat\ff^*(x_i) ~~~~~\text{where}~~~~~~ \hat \ff^*(x_i) = \hat \E_{x\in \{x_i\}} [ k(x_i, ~ x) \nabla_{x} \log p(x)  + \nabla_{x} k(x_i, x) ], 
&&&z_i ~ \gets ~ z_i  ~  + ~ \epsilon \ff^*(z_i), 
~~~~~~~~\forall i = 1, \ldots, n,  
\\
\!\!\!\!\!\!\!
&&&\ff^*(z_i) = \frac{1}{n} \sum_{j=1}^n [  \nabla_{z_j} \log p(z_j) k(z_j, z_i) + \nabla_{z_j} k(z_j, z_i) ]. \notag
%x_i ~ \gets ~ x_i  ~  + ~ \epsilon \frac{1}{n} \sum_{x\in \{x_i\}} [ k(x_i, ~ x_j) \nabla_{x} \log p(x)  + \nabla_{x} k(x_i, x) ] . 
%\end{split}
\end{align}
%and $\hat\E_{x\sim \{x_i\}_{i=1}^n}[f(x)] = \sum_i f(x_i)/n$. 
The two terms in $\ff^*(z_i)$ play two different roles: 
the term with the gradient $\nabla_z \log p(z)$ drives the particles toward the high probability regions of $p(z)$, 
while the term with $\nabla_z k(z,z_i)$ serves as a repulsive force to encourage different particles to be different from each other as shown in \citet{liu2016stein}. 
%to see this, assume we use a stationary kernel $k(x,x') = k(x-x')$, then the second term reduces to $\hat \E_x \nabla_{x} k(x,x_i) = - \hat \E_x \nabla_{x_i} k(x,x_i)$  ($\hat \E$ denotes the empirical averaging on $\{x_i\}$), which can be treated as the negative gradient for minimizing the average similarity $\hat \E_x k(x,x_i)$ in terms of $x_i$. 
Overall, this procedure provides diverse points for approximating distribution $p$ when it converges. 
%and uncertainty assessment. 
%, and also has an interesting ``momentum'' effect in which the particles move collaboratively to escape the local optima.
%See \citet{liu2016stein} for more details. 

It is easy to see from \eqref{equ:update11} that $\ff^*(z_i)$ reduces to the typical gradient $\nabla_z \log p(z_i)$ when there is only a single particle ($n=1$) and $\nabla_z k(z,z_i) = 0$ when $z=z_i$,  
in which case SVGD reduces to the standard gradient ascent for maximizing $\log p(z)$ (i.e., maximum \emph{a posteriori} (MAP)).

%Substinting 
\paragraph{Computing the Kernelized Stein Discrepancy}

%One can obtain a computational tractable form for the kernelized Stein discrepancy (KSD).  
%
By substituting the $\ff^*$ in \eqref{equ:ffss} into \eqref{equ:ksd},  
one can show that \citep{liu2016kernelized, chwialkowski2016kernel, oates2014control} %showed that 
%Kernelized Stein discrepancy %\citep{liu2016kernelized, chwialkowski2016kernel, oates2014control} (KSD)
%\citep{liu2016kernelized} (KSD)
%is a discrepancy statistic for measuring differences between two probability distributions based on Stein's identity, which has the form:
\begin{align}
\label{equ:ksdexp} 
{\S(q  ~||~ p)}^2 =  \E_{\z, \z' \sim ~ q} [\kappa_p (\z,\z')] , 
%\\& \text{where} ~~~~ \stein_p f(\z) = \E_{\z\sim q}[\score_p(\z) f(\z)  + \nabla_\z f(\z)], 
\end{align}
where $\kappa_p(\z,\z')$ is a positive definite kernel obtained by applying Stein operator on $k(\z,\z')$ twice, as a function of $z$ and $z'$, respectively. 
%which has a computationally tractable form: 
%Importantly, $\kappa_p(\z, \z')$ 
It has the following computationally tractable form: 
\begin{align*}
\kappa_p(\z, \z') 
= &\score_p(\z)^\top k(\z,\z') \score_p(\z') + \score_p(\z)^\top \nabla_{\z'} k(\z, \z') \notag \\ 
&+\nabla_\z k(\z,\z')^\top \score_p(\z') + \nabla_\z^\top (\nabla_{\z'} k(\z,\z')),  
\end{align*}
where $\score_p(\z) = \nabla_\z\log p(\z)$.
%It can be shown that $\S(q~||~p)=0$ if and only if $q = p$ when $k(\x,\x')$ is strictly positive definite in a proper sense \citep{liu2016kernelized, chwialkowski2016kernel}.
The form of KSD in \eqref{equ:ksdexp} provides a computationally tractable way for estimating the Stein 
 discrepancy between a set of samples $\{\z_i\}$ (e.g., drawn from an unknown $q$) and a distribution $p$ specified by its score function $\nabla_\z\log p(\z)$ (which is independent of its normalization constant), %using $U$-statistics which unbiasedly estimates KSD: 
\begin{align}\label{equ:uv}
\hat \S_u^2(\{\z_i\}~||~p) =  \frac{1}{n(n-1)}\sum_{i\neq j} [\kappa_p(\z_i, \z_j)],
% \hat \S_v^2(\{\z_i\}~||~p) =  \frac{1}{n^2}\sum_{ i,j} [\kappa_p(\z_i, \z_j)], 
\end{align}
where $\hat \S_u^2(q~||~p)$ provides an unbiased estimator (hence called a $U$-statistic) for $\S^2(q~||~p)$.
As a side result of this work, 
we will discuss the possibility of using KSD as an objective function for wild variational inference in Appendix~\ref{sec:ksd}. % and Section~\ref{para:gmm}. 

%This update evolves the particles collectively to match the target distribution with the two terms of $\Delta(x_i)$ playing two different roles:    the first term in $\Delta(x_i)$ is theweighted sum of the particles' gradient of the log-density, which drives the particles towards the high probability regions in a collaborative fashion; the second term can be shown to serve as a ``repulsive force'' that makes the particles repel each other to maintain a degree of diversity.
% maintaining a degree of diversity via the second term $\nabla_x k(x_i, x)$ that makes the particles repel each other.

\begin{algorithm}[t]                      % enter the algorithm environment
\caption{Amortized SVGD for Problem~\ref{pro:prob1}}% for Wild Variational Inference}          % give the algorithm a caption
\label{alg:alg1}                           % and a label for \ref{} commands later in the document
\begin{algorithmic}                    % enter the algorithmic environment
\STATE Set batch size $m$, step-size scheme $\{\epsilon_t\}$ and kernel $k(z,z')$. Initialize $\eta^0$. 
\FOR {iteration $t$}
\STATE  Draw random $\{\xi_i\}_{i=1}^m$, calculate $z_i = f(\eta^t;~\xi_i)$, 
and the Stein variational gradient $\ff^*(z_i)$ in \eqref{equ:update11}. 
%\begin{align*}%\label{equ:dxi}
%\Delta x_i = \hat\E_{x\sim\{x_i\}_{i=1}^m} [\log p(x) k(x,x_i) + \nabla_x k(x, x_i)]. 
%\end{align*}
\STATE  Update parameter $\eta$ using either \eqref{equ:follow1}, \eqref{equ:follow2} or \eqref{equ:follow3}. 
%$$\eta^{t+1} \gets \eta^{t} + \epsilon ~ \sum_{i=1}^n \partial_\eta f(\eta^t;~\xi_i)  \Delta x_i.$$ 
\ENDFOR
\end{algorithmic}
\end{algorithm}

%!TEX root = ../main.tex
\section{AMORTIZED SVGD: TOWARDS AN AUTOMATIC NEURAL SAMPLER}
\label{sec:amortizedsvgd}
%\section{Our Method}

%SVGD and other particle-based methods require a large memory to store the particles when  using a large number particles. In addition, 
%Particle methods like SVGD can provide simple and consistent approximation for individual target distributions, 
SVGD and other particle-based methods become inefficient when we need to apply them repeatedly on 
a large number of different, but similar target distributions for multiple tasks, %including online learning or inner loops of other algorithms, 
because they can not leverage the similarity between the different distributions 
%adaptively improve themselves based on the experience from the past tasks, 
and may require a large memory to restore a large number of particles. 
This problem can be addressed by training a neural network $f(\xi;~\eta)$ to output particles that would 
have been produced by SVGD; %(or fixed points of the SVGD dynamics); 
this amounts to  ``amortizing'' or compressing the nonparametric SVGD 
into a parametric network, 
%We propose to ``amortize'' the SVGD dynamics 
% by training a neural network $f(\xi;~\eta)$ to 
% output samples that satisfies the fixed point of SVGD, 
yielding a solution for wild variational inference in Problem~\ref{pro:prob1} of Section~1. %%mimic the SVGD dynamics, 

%We propose to use a neural network $f_{\eta}$ so that the output $x = f(\eta;~\xi)$ with $\xi \sim q_0$ mimics the SVGD dynamics. 
One straightforward way to achieve this is to run SVGD until convergence and 
train $f(\xi;~\eta)$ to fit the resulting SVGD particles (e.g., by using generative adversarial networks (GAN) \citep{goodfellow2014generative}). 
This, however, requires running many epochs of fully converged SVGD and can be slow in practice. 
We instead propose an \emph{incremental approach} in which $\eta$ is iteratively adjusted 
so that the network outputs $z = f(\xi;~\eta)$ improves by moving along the Stein variational gradient direction in \eqref{equ:update11}, 
in order to move towards the target distribution. 
%decrease the KL divergence between the target and the approximation distribution. 
%to follow the change $\ff^*(x_i)$ given by SVGD. 
% to mimic the SVGD updates which can summary 
%the experience from the past tasks concisely, allowing ``learning to draw samples''. 
%In this 
%Many practice applications involve a large number of target distributions with similar structures for which 
%it is inefficient to repeatedly applying SVGD or other tradition inference methods on the individual distributions separately. 
%and it would be highly desirable to intelligent systems that \emph{train itself to make better Bayesian inference over time, and also generalize well to new tasks with similar structures}. 
%becomes inefficient when we have a large number of different target distributions that requires us to run the particle algorithm repeatedly. 
%The different target
%This happens when we need to reason with different observed data or priors, or need to make iterative, real-time inference for online tasks or inner loops of learning algorithms. 
%Therefore, it is highly desirable to develop more intelligent Bayesian inference systems that can adaptively improve its own performance over time, and also generalize well to new tasks with similar structures. 

%Our idea is simple: 
%we want to iteratively adjust $\eta$ such that the black-box output $x = f(\eta;~D, \xi)$ moves along the Stein variational gradient direction $\Delta(x_i)$ in \eqref{equ:update11} in order to decrease the KL divergence between the target and approximation distribution. 
Specifically, denote by $\eta^t$ the parameter estimated at the $t$-th iteration of our method; 
each iteration of our method 
draws a batch of random inputs $\{\xi_i\}_{i=1}^m$ 
and calculate their corresponding output $z_i = f(\xi_i;~\eta^t)$ based on $\eta^t$, where $m$ is a mini-batch size (e.g., $m=100$). 
%as well as $x'_i  = x_i \epsilon \ff^*(x_i)$%B
The Stein variational gradient $\ff^*(z_i)$ in 
\eqref{equ:update11} would then ensure that $z'_i = z_i + \epsilon \ff^*( z_i)$ forms a better approximation of the target distribution $p$. 
Therefore, we should adjust $\eta$ to make it output $\{z'_i\}$ instead of $\{z_i\}$, 
that is, we want to update $\eta$ by
\begin{align}\label{equ:follow1}
\eta^{t+1} \gets & \argmin_\eta  \sum_{i=1}^m || f(\xi_i;~\eta)  - z_i' ||_2^2,
\end{align}
where $z_i'  = z_i  + \epsilon \ff^*(z_i)$. 
This process is repeated until convergence, 
in which case the outputs of network $f$ %should be the fixed points of
can no longer be improved by SVGD and hence should form a good approximation of the target $p$. 
See Algorithm~\ref{alg:alg1}. % for the summary of this procedure. 

If we assume $\epsilon$ is very small, then 
\eqref{equ:follow1} can be approximated by a least square optimization. To see this, note that 
$f(\xi_i;~\eta) \approx f(\xi_i;~\eta^t) + \partial_\eta f(\xi_i;~\eta^t) (\eta - \eta^t)$ by Taylor expansion. 
Since $z_i = f(\xi_i;~\eta^t)$, we have
%Therefore, 
$$
 ||  f(\xi_i;~\eta)  - z_i' ||_2^2 \approx || \partial_\eta f(\xi_i;~\eta^t) (\eta  - \eta^t)  - \epsilon \ff^*(z_i)  ||_2^2. 
$$
%This allows us to approximate the update i
As a result, \eqref{equ:follow1} reduces to a least square optimization: 
\begin{align}\label{equ:follow2}
\begin{split}
&  \eta^{t+1} \gets  \eta^t - \epsilon \Delta \eta^t, \\ 
%\text{where~~}
& \Delta \eta^t = \argmin_{\delta}  \sum_{i=1}^m   || \partial_\eta f(\xi_i;~\eta^t)  \delta   -  \ff^*( z_i ) ||_2^2.
\end{split}
\end{align}
%Although $\Delta \eta^t$ can be solved with a least square optimization, 
It may be still slow to solve a least square problem at each iteration. 
%Update \eqref{equ:follow2} can still be computationally expensive because of the need for %matrix inversion in
%solving the least square problem. 
% in each iteration. 
We can derive a more computationally efficient approximation by performing only one step of gradient descent of \eqref{equ:follow1} starting at $\eta^t$ (or equivalently \eqref{equ:follow2} starting at $\delta = 0$), which gives 
\begin{align}\label{equ:follow3}
\eta^{t+1} \gets \eta^t + \epsilon  \sum_{i=1}^m   \partial_\eta f(\xi_i;~\eta^t) \ff^*(z_i). 
\end{align}
%based on update \eqref{equ:follow3}. 
%We find this update works efficiently in practice. It is shown in Algorithm~\ref{alg:alg1}
%
%In this way, everything is nice
Although update \eqref{equ:follow3} is derived as an approximation of \eqref{equ:follow1} or \eqref{equ:follow2}, 
it is computationally faster and it works effectively in practice; 
this is because when $\epsilon$ is small, one step of gradient update can be sufficiently close to the optimum.  

%Update \eqref{equ:follow3} has a particularly 
%Intuitively speaking, 
Update \eqref{equ:follow3} has a simple and intuitive interpretation: it can be thought as \emph{a ``chain rule'' that back-propagates the Stein variational gradient to the network parameter $\eta$}. 
%making the random sample generated by $z = f(\eta;~\xi)$ mimic the behavior of SVGD and hence move towards the target distribution. 
%Although $\ff^*(x_i)$ is not a gradient in the typical sense, it can be interpreted as a form of functional gradient with which the update of $\eta$ can be justified theoretically. 
This can be justified by considering the special case when we use only a single particle $(n=1)$ 
in which case $\ff^*(z_i)$ in \eqref{equ:update11} reduces to the typical gradient $\nabla_z \log p(z_i)$, %of $\log p(x)$, 
%In particular, note that if we just use a single particle for each $D$, our method reduces 
and update \eqref{equ:follow3} reduces to the typical gradient ascent for maximizing $\E_{\xi}[\log p(f(\xi;~\eta))],$ 
 in which case $f(\xi;~\eta)$ is trained to maximize $\log p(z)$ (that is, \emph{learning to optimize}), 
instead of \emph{learning to draw samples from $p$} for which it is crucial to use the Stein variational gradient $\ff^*( z_i)$ to diversify the outputs to capture the uncertainties in $p$. 
%instead of drawing sampling from $p$ which requires to diversify the network output. 
%the method by \citet{andrychowicz2016learning} for learning to optimize using gradient descent.

Update \eqref{equ:follow3} also has a close connection with the typical variational inference with the reparameterization trick \citep{kingma2013auto}. Let $q_\eta(z)$ be the density function of $z = f(\xi;~\eta)$, $\xi\sim q_0$. Using the reparameterization trick, the gradient of $\KL(q_\eta~||~p)$ w.r.t. $\eta$ equals % can be shown to be 
\begin{align*}
\nabla_\eta\KL(q_{\eta}~||~p) 
= -\E_{\xi \sim q_0}[ \partial_\eta f(\eta;~\xi) %\\
	 \nabla_z \log (p(z)/q_\eta(z))]. 
\end{align*}
With $\{\xi_i\}$ i.i.d. drawn from $q_0$ and $z_i = f(\xi_i; ~\eta), ~\forall i$, we can obtain a standard stochastic gradient descent for minimizing the KL divergence:  
%can be performed by 
\begin{align}\label{equ:rep}
\begin{split}
&\eta^{t+1} \gets \eta^t +  \sum_{i=1}^m \partial_\eta f(\xi_i;~\eta^t) \tilde \ff^*(z_i), \\
\text{where} ~~~~ 
&\tilde \ff^*(z_i) = \nabla_z \log p(z_i) - \nabla_z \log q_{\eta^t}(z_i). 
\end{split}
\end{align}
This is similar to \eqref{equ:follow3}, but replaces the Stein gradient $\ff^*(z_i)$ defined in \eqref{equ:update11} with 
$\tilde \ff^*(z_i)$. 
However, because $\tilde \ff^*(z_i)$ depends on the density $q_{\eta^t}$, 
which is assumed to be intractable in Problem 1, \eqref{equ:rep} is not directly applicable in our setting. 
%The advantage of using $\ff^*(x_i)$ is that it does not require to explicitly calculate $q_\eta$, and hence admits a solution to Problem 1 in which $q_\eta$ is not computable for complex network $f(\xi;~\eta)$ and unknown input distribution $q_0$. 
Further insights can be obtained by noting that 
\begin{align}
\label{equ:tmp}
\ff^*(z_i) 
& \approx \E_{z\sim q_{\eta^t}}[\nabla_z \log p(z)k(z,z_i) + \nabla_zk(z,z_i)] \notag \\
& =  \E_{z\sim q_{\eta^t}}[(\nabla_z \log p(z) - \nabla_z \log q_{\eta^t}(z))k(z,z_i)]   \notag \\
& = \E_{z\sim q_{\eta^t}} [ \tilde \ff^*( z) k(z, z_i)],  %& = \E_{x\sim q} [k(x, x_i) \tilde \Delta x]  \notag
%& = \frac{1}{n}\sum_{j= 1}^n [ \tilde \ff^*( x_j) k(x_j, x_i)],  %& = \E_{x\sim q} [k(x, x_i) \tilde \Delta x]  \notag
%\\&\approx  \sum_j k(x_j, x_i) \tilde \ff^*(x_i) \notag
\end{align}
where \eqref{equ:tmp} is obtained by using Stein's identity \eqref{equ:steinid}. 
%Therefore, $\ff^*(x_i)$ is approximately $\tilde \ff^*(x_i)$ multiplied by a positive definite matrix $[k(x_i, x_j)]_{ij}$ and hence with positive inner product.% $\ff^*(x_i), \tilde \ff^*(x_i)\la $. 
Therefore, $\ff^*(z_i)$ can be treated as a smoothed version of $\tilde \ff^*(z_i) $ obtained by convolving it with kernel $k(z,z')$. 
%It is also possible to get $q_\eta(x)$-free (wild) variational inference by directly approximating $\nabla_x\log q_\eta(x)$ based on 

%!TEX root = ../main.tex
\section{APPLICATIONS OF AMORTIZED SVGD}\label{sec:apps}

With amortized SVGD, we can
use expressive inference networks to obtain better approximation and 
 explore new applications where traditional VI methods cannot be applied.  
In this section, we introduce two different applications of amortized SVGD. 
One is  training variational autoencoders \citep{kingma2013auto} 
with complex, non-Gaussian encoders, and the other is  
training ``smart'' MCMC samplers that adaptively improve their own hyper-parameters from past experience. 
%when performed a group of similar distributions.  
%\red{black-box} samplers that can draw samples from arbitrary distributions from a well defined family distribution.

\subsection{Amortized SVGD For Variational Autoencoders}\label{sec:vae}
% \todo{x, z are not consistent with the other places, use z instead of x in other places?}
% \todo{use $\eta$ instead of $\eta$ to consistent with other sections}
Variational autoencoders (VAEs) \citep{kingma2013auto} are latent variable models of form
 $p_\theta(x) = \int_z p_\theta(x | \z) p_\theta(z) dz$ where 
 $\x$ is an observed variable and $\z$ is an un-observed latent variable. 
 Assume the empirical distribution of the observed variable is $\hat p(x)$, 
VAE learns the parameter $\theta$ using a variational EM algorithm 
which approximates the posterior distribution $p_\theta (z | x)$ with a simple encoder $q_\eta(z | x)$, and updates $\theta$ and $\eta$ alternatively by 
% to train $\theta$ and posterior distribution $q_\eta(z | x)$: 
%  of the visible variables $\x$ given the latent variables $\z$, with a prior $p_\theta(\bf z)$ over the latent variables, and an approximate inference model $q_\eta ( z \mid x)$ over the latent variables given the visible variables. 
%The training process of VAEs can be viewed as performing Variational EM algorithm, where E-step is to update the encoder parameter $\eta$:
\begin{align}
& \theta \gets \argmax_{\theta}~ \E_{\hat p(x)q_\eta(z \mid x)}\big[\log p_\theta(x, ~ z)\big], \label{equ:m}\\
& \eta \gets \argmin_{\eta}~ 
\E_{\hat p(x)}\big[\mathrm{KL}(q_\eta(z \mid x) ~|| ~p_\theta(z \mid x))\big],  \label{equ:evae}
\end{align}
which alternates between updating $\theta$ by maximizing the joint likelihood (M-step \eqref{equ:m}) 
and approximating the posterior distribution $p_\theta (z|\x)$ given fixed $\theta$ with  variational inference (VI) (E-step \eqref{equ:evae}). 
%and M-step is to update the decoder parameter $\theta$:
%\begin{align*}
%\end{align*}
%Due to the intractability of the posterior $p_\theta(z \mid ~x)$, 
%When performing E-step in equation \eqref{equ:evae}, 
In standard VAE,
 \eqref{equ:m} is performed using standard VI with the reparameterization trick \eqref{equ:rep},
 which requires $q_\eta$ to be tractable. 
Therefore, $q_\eta(z \mid x)$ is often defined as a Gaussian distribution with mean and diagonal covariance parameterized by neural networks with $x$ as input. 
%As observed in \citet{adversial bayesian,  normalizing flow, auteregress flow}, %improvement can be obtained by re
This Gaussian assumption potentially limits the quality of the resulting generative models,
and more expressive encoders can improve the 
performance as shown in recent works \citep[e.g,][to name a few]{
 kingma2016improving, mescheder2017adversarial}.
 %This can be too simple to capture the 
%While the inference model is very expressive to capture the information of $x$, the restrictive assumption on $z$ potentially limits the quality of the resulting generative model.

By applying amortized SVGD to solve the posterior inference problem in \eqref{equ:evae}, %or \eqref{equ:entropy_vae}, 
we obtain simple algorithms that work with more complex inference networks. 
Specifically, we assume that $z \sim q_\eta(z|x)$ 
is generated by $z = f(\xi, x; ~ \eta)$, and optimize $\eta$ using update 
\eqref{equ:follow1}-\eqref{equ:follow3}. 
See Algorithm~\ref{alg:alg2}. 
In our experiment, we take $z = f(\xi, x; ~\eta)$
to be a deep neural network with binary Bernoulli dropout noise at the input layer of the network
which is more effective in approximating multi-modal posteriors than the simple Gaussian encoders.

% for solving \eqref{equ:evae}.
%To solve \eqref{equ:entropy_vae}, we just need to multiply constant $\alpha + 1$ on the second term in equation \eqref{equ:update11}. Detail explanation can be found in the appendix. 
%\red{
Similar idea has also been explored in \citet{punips17steinvae}. Besides, we propose an \emph{entropy regularized VAE} 
%that can further improve the performance. 
to improve the standard VAE and get more diverse images by adding an entropy regularization term on the encoder networks. 
Here, we replace the $\eta$-update in \eqref{equ:evae} with 
%we can iteratively update $\eta$ by 
\begin{align}
% [ \mathrm{KL}(&q_\eta(z \mid x) ~|| ~p_\theta(z \mid x))\big] 
\eta 
& \gets \argmin_{\eta} \E_{\hat p(x)} [ \mathrm{KL}(q_\eta(z \mid x) ~|| ~p_\theta(z \mid x))\big] 
- \alpha H(q_\eta) \notag \\
& = \argmin_{\eta} \E_{\hat p(x)}  \big[ \mathrm{KL}(q_\eta(z \mid x) ~|| ~ p_\theta(z \mid x)^{\frac{1}{1+\alpha}})\big],   
%\notag \\  &+ \alpha\E_{q_\eta(z |x)}[\log q [ \mathrm{KL}(&q_\eta(z \mid x) ~|| ~p_\theta(z \mid x))\big] _\eta(z |x)], 
\label{equ:entropy_vae}
\end{align}
where $H(q_\eta) =  - \E_{\hat p(x)q_\eta(z |x)}[\log q_\eta(z |x)]$ is the (conditional) 
entropy of $q_\eta(z | x)$, and $\alpha$ is a regularization coefficient. 
\eqref{equ:entropy_vae} can be solved by 
applying SVGD on the tempered distribution $p_\theta(z \mid x)^{\frac{1}{1+\alpha}}$, 
which can be done by $z_i ~ \gets ~ z_i  ~  + ~ \epsilon \ff^*(z_i)$ with 
\begin{align}\label{equ:update11alpha}
%&&&z_i ~ \gets ~ z_i  ~  + ~ \epsilon \ff^*(z_i), ~~~~~~~~\forall i = 1, \ldots, n,  
%\\
%\!\!\!\!\!\!\!
%&&&
\begin{split}
\!\!\!\!\!\!\!
\ff^*(z_i) = \frac{1}{n} \sum_{j=1}^n [  &  \nabla_{z_j} \log p_\theta(z_j | x) k(z_j, z_i) ~~+  \\
& ~~~~~~~~~~~~~~~~~+ ~~ (1+\alpha) \nabla_{z_j} k(z_j, z_i) ], 
\end{split}
%\notag
\end{align}
where the temperature parameter $(1+\alpha)$ 
becomes a weight coefficient of the repulsive force; 
a high temperature (or equivalent a large entropy regularization) yields 
a strong repulsive force and push the particles to be further away from each other. 
%which yields update $z_i \gets z_i + \epsilon \phi$
%}

% I comment following sentences
\begin{comment}
%With amortized SVGD,
By applying amortized SVGD to solve the posterior inference problem in \eqref{equ:evae}, 
we obtain a simple algorithm that work with more complex inference networks. 
Specifically, we assume that $z \sim q_\eta(z|x)$ 
is generated by $z = f(\xi, x; ~ \eta)$, and optimize $\eta$ using update 
 \eqref{equ:follow1}-\eqref{equ:fnollow3}. See Algorithm~\ref{alg:alg2}.
\end{comment}

\begin{algorithm}[t]                      % enter the algorithm environment
\caption{Amortized SVGD for training VAE}% for Wild Variational Inference}          % give the algorithm a caption
\label{alg:alg2}                           % and a label for \ref{} commands later in the document
\begin{algorithmic}                    % enter the algorithmic environment
\STATE Set batch size $m$ and kernel $k(\z, \z')$. 
\STATE Initialize encoder parameter $\eta$ and decoder parameter $\theta$.
\FOR {iteration $t$}
\STATE Pick input $x^{(i)}$ from the training data.
\STATE Draw $\{\xi_k\}_{k=1}^m\sim q_0$, calculate $z_k^{(i)}=f(x^{(i)}, \xi_k;~\eta)\vspace{.1cm}$, and their Stein variational gradients $\ff^*(z_k^{(i)})$ in \eqref{equ:update11} or \eqref{equ:update11alpha} 
related to gradient $\nabla_z \log p_\theta(z_k^{(i)}|x^{(i)})$. %\red{XXXXX}
\STATE Update $\eta$ using either \eqref{equ:follow1}, \eqref{equ:follow2} or \eqref{equ:follow3}. 
\STATE Update $\theta$ with \\ $~~~~~~~~~ \theta \gets \theta +  \frac{1}{m} \sum_{k=1}^{m}\nabla_\theta \log p_\theta(x^{(i)}, ~ z_k^{(i)})$.
\ENDFOR
\end{algorithmic}
\end{algorithm}

\subsection{Training Langevin Samplers}
\label{sec:trainlangevin}
By viewing typical MCMC procedures as a simulator $f(\cdot)$, 
we can apply amortized SVGD to adaptively improve hyperparameters in MCMC inference. 
This is useful when we need to 
perform Bayesian inference on a large number of different, but similar datasets or posteriors,
where we can adaptively improve the MCMC sampler for future tasks by leveraging the information of the previous tasks.  
An example of this, which we consider in this work, is adaptively learning optimal step sizes for Langevin dynamics. 
%\red{As an example, we consider learning step size of Langevin dynamics as an illustration of our methology. }

%We consider s
%allowing us to leverage similarThis allows us to significantly improve the computation efficiency when we need to perform Bayesian inference on a large number of different tasks. 
%
To specify the general framework, %problem, 
we assume that we are interested in drawing samples from a set of distributions 
$$\Q = \{ p_\vartheta(\z)  \colon \vartheta \in \Theta\},$$ 
indexed by parameter $\vartheta$. 
We are interested in learning 
a network $f(\xi, p_\vartheta; ~\eta)$ 
which maps the distribution $p_\vartheta$ to stochastic posterior samples. 
Note that this is a generalization of Problem 1 which 
focuses on approximating an individual distribution. 
In practice, $p_\vartheta$ could be the posterior distributions of unknown parameters of interest conditioning on different observed data, 
or models of different individuals in hierarchical models.  

In order to learn the network $f(\xi, p_\vartheta; ~\eta)$, we modify Algorithm~\ref{alg:alg1}, 
to perform amortized SVGD on a randomly selected $p_\vartheta$ from $\Q$ at each iteration.  
See Algorithm~\ref{alg:alg3}. 
In this way, we expect that the trained network $f(\xi, p_\vartheta; ~\eta)$ can perform well on similar $p_\vartheta$ drawn from the same distribution, 
but never seen by the training algorithm.
%but drawn from the same distribution as the training algorithm sees. 
%where w

A useful perspective is that typical MCMC methods can be viewed 
as neural networks $f(\xi, p_\vartheta; ~\eta)$ 
handcrafted by researchers, with nice theoretical properties. 
We can leverage the structure of existing MCMC to 
design the architecture of $f(\cdot)$, 
and use amortized SVGD to adaptively improve its parameters across different tasks. 

% We take Langevin dynamics 
 As an example, 
Langevin dynamics draws samples from $p_\vartheta$ 
by starting with an initial sample $z^0$ and performing iterative random updates of form  
$z^{t+1} \gets f_t(z^t)$ with 
\begin{align}\label{equ:langevin} 
%& z^{t+1} \gets f_t(z^t), \\
& f_t(z^t) = \z^{t} + \eta^t \odot  \nabla_\z \log p_\vartheta(\z^t)   + \sqrt{2\eta^t} \odot \xi^t, 
\end{align}
which performs gradient ascent with a Gaussian perturbation.  
Here $\eta^t$ denotes a vector-valued step size at the $t$-th iteration and ``$\odot$'' denotes element-wise product, 
and $\xi^t$ is a standard Gaussian random vector of the same size as $\z^{t}$.

\begin{algorithm}[tbp]                      % enter the algorithm environment
\caption{Amortized SVGD for Learning to Sample a distribution family $\Q = \{p_\vartheta \colon \vartheta \in \Theta \}$}%for Problem \ref{pro:prob2}}% for Wild Variational Inference}          % give the algorithm a caption
\label{alg:alg3}                           % and a label for \ref{} commands later in the document
\begin{algorithmic}                    % enter the algorithmic environment
\STATE Goal: Learn parameter $\eta$ to train $z = f(\xi, p_\vartheta; ~\eta)$ drawing samples from $p_\vartheta$. 
\STATE Set batch size $m$, and kernel $k(\z,\z')$. Initialize $\eta^0$.  
\FOR {iteration $t$}
\STATE Randomly select $p_\vartheta$ from $\Q$.
\STATE Draw random seed $\{\xi\}_{i=1}^{m}\sim q_0.$
\STATE Calculate $z_i = f(\xi_i, ~ p_\vartheta; ~\eta)$, 
and the Stein variational gradient $\ff^*(z_i)$ in \eqref{equ:update11}. 
\STATE  Update parameter $\eta$ using either \eqref{equ:follow1}, \eqref{equ:follow2} or \eqref{equ:follow3}. 
\ENDFOR
\end{algorithmic}
\end{algorithm}

We can view $T$ iterations of Langevin dynamics as a $T$-layer neural network: 
\begin{align}\label{equ:ls} 
z^T = f(\xi, ~ p_\vartheta; ~ \eta) = f_{T-1} \circ \cdots \circ f_0(\xi, ~ p_\vartheta; ~ \eta), 
\end{align}
in which 
the initial samples and the Gaussian noise form the random seeds of the network, that is, $\xi =\{\xi^t\}_{t = 0}^{T-1}\cup \{z^0\}$, and 
the step sizes $\eta=\{\eta^t\}_{t = 0}^{T-1}\vspace{0.1 cm}$ form the parameters which we can estimate using Algorithm~\ref{alg:alg3}.  %$% we want estimate. 
In cases when 
it is difficult to calculate $\nabla_z \log p_\vartheta$ exactly, such as the case of Bayesian inference with 
big datasets, 
we can use stochastic gradient Langevin dynamics \citep{welling2011bayesian}
to approximate $\nabla_z \log p_\vartheta$ with subsampling, which introduces another source of randomness. 
This, however, does not influence the application of Algorithm~\ref{alg:alg3}, 
since our method does not need to know the structure of the random seed distribution. 
%only need to differentiate through $\eta$, regardless 
%We demonstrate the  of  several  examples. 
 %$\xi =\{\xi^t\}_{t = 0}^{T-1}\cup \{z^0\}$ denotes the randomness the network, and $\theta$ is the parameter that specify the target distribution $p$. 
%For simplicity, we refer this structure of $f(\xi, ~ p(\z; ~\theta); ~\eta)$ as \textit{Langevin sampler}. 
%We demonstrate several  examples of problem \ref{pro:prob2} utilizing Langevin sampler structure, which can be found in section \ref{sec:sampler}.

In practice, a large value of $T$ would result in 
a deep network and cause a vanishing gradient problem. 
We address this problem by partitioning the $T$ layers into small blocks of size $5$ or $10$, 
and evaluate the gradient of the parameters in each block by back-propagating the Stein variational gradient from the output of its own block.

\newcommand{\btt}{\begin{tabular}{c}}
\newcommand{\ett}{\end{tabular}}
\newcommand{\fttmp}{\scriptsize}

\begin{figure*}[t]
\centering
\begin{tabular}{ccccc}
\includegraphics[width=0.16\textwidth]{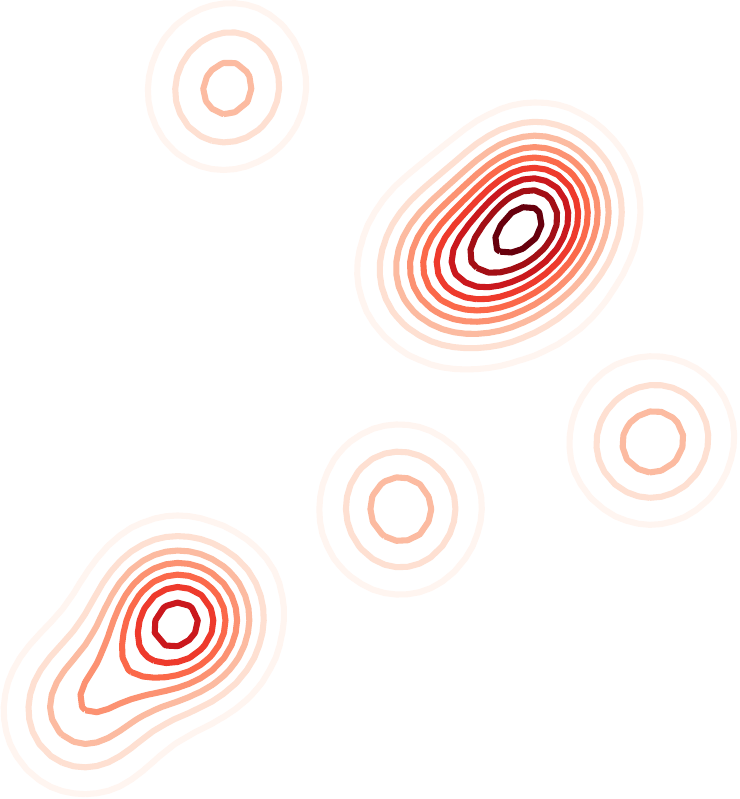} &
\includegraphics[width=0.15\textwidth]{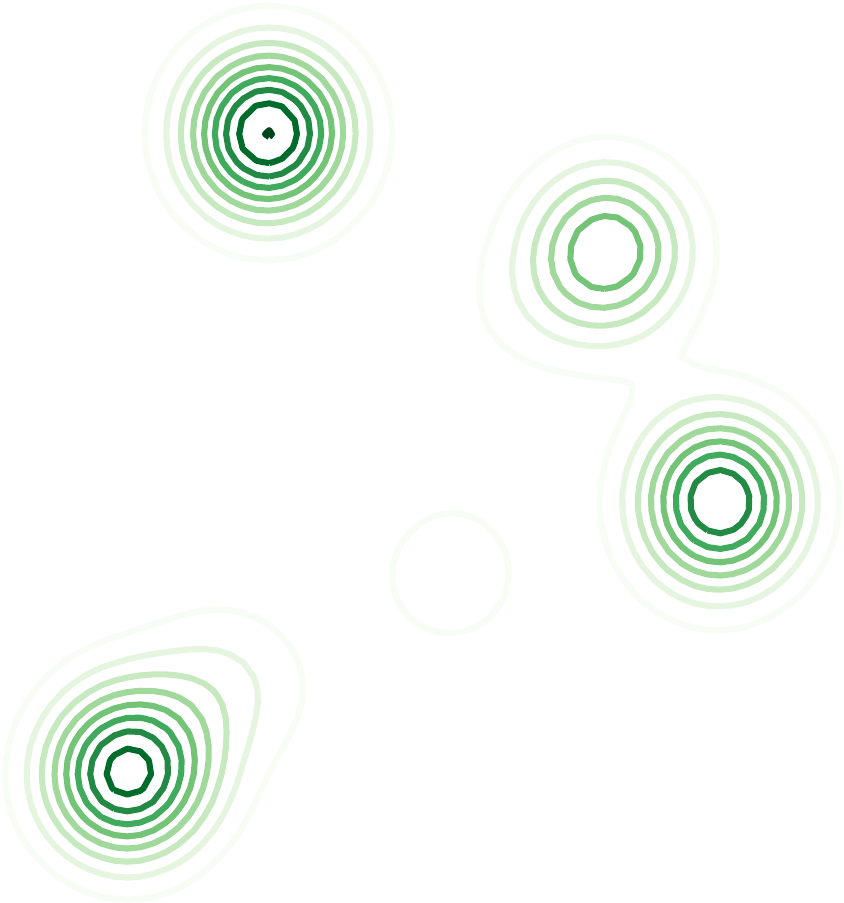} &
\includegraphics[width=0.16\textwidth]{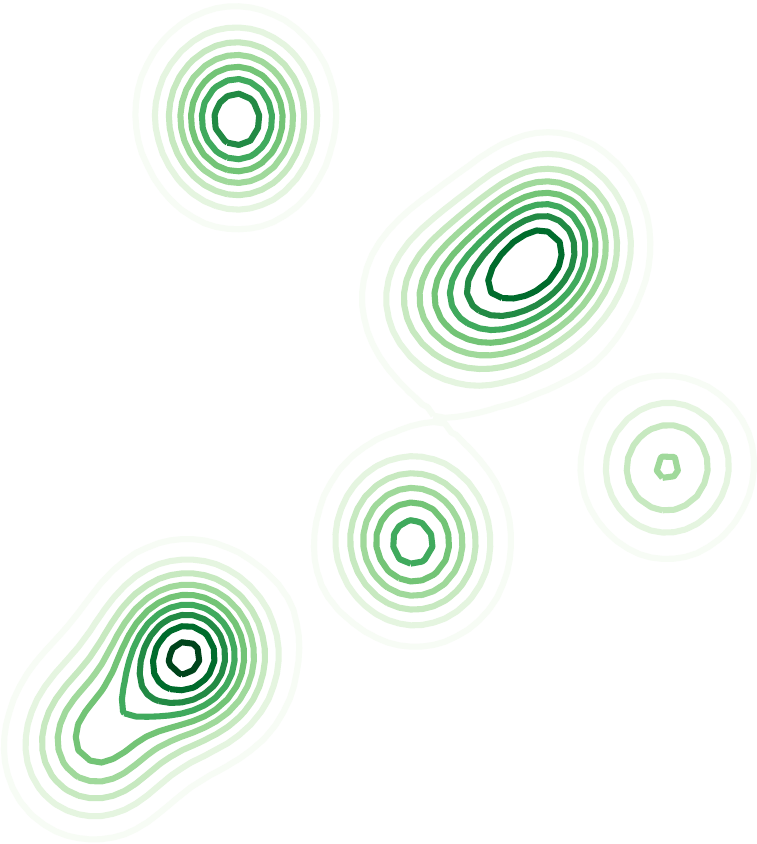} &
\includegraphics[width=0.16\textwidth]{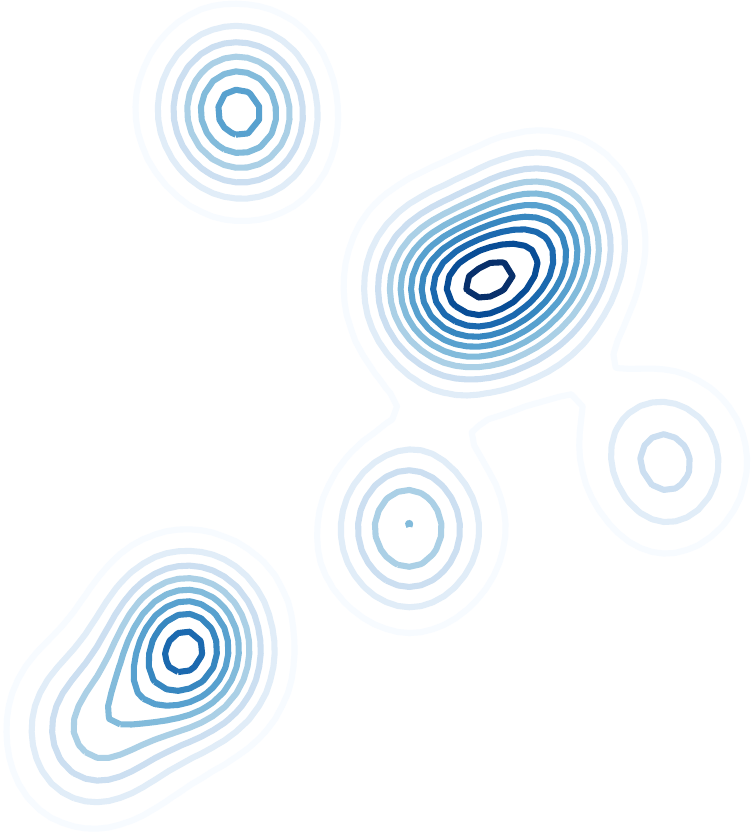} &
\includegraphics[width=0.16\textwidth]{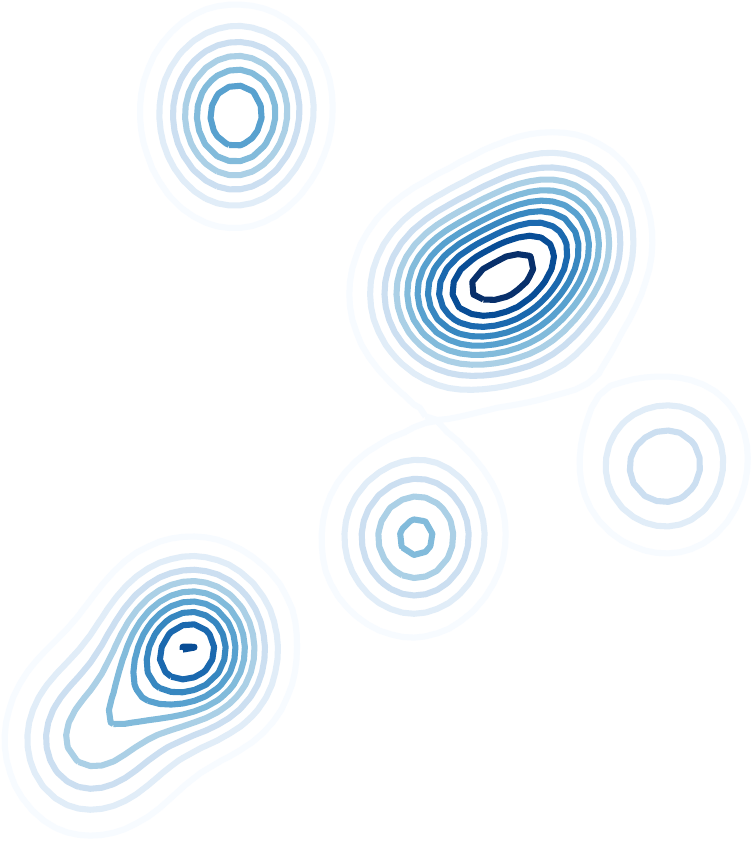} \\
\fttmp \btt (a) Test Distribution \\(truth) \ett &\fttmp \btt (b) Power Decay \\ (15 steps) \ett &
 \fttmp \btt (c) Power Decay\\ (1,000 steps) \ett &
 \fttmp \btt (d) Amortized SVGD  \\ (5 steps)  \ett 
  & \fttmp \btt (e)  Amortized SVGD\\ (15 steps) \ett 
%  & \fttmp \btt (f) Amortized KSD \\ (15 steps) \ett
\end{tabular}
\vspace{-10bp}\caption{\small 
Learning to sample from GMM. We train Langevin samplers using amortized SVGD
%or amortized KSD \red{(introduced in Appendix \ref{sec:ksd})}
on a set of randomly generated GMMs, and 
evaluate the samplers on a new GMM shown in (a) 
generated randomly in the same way as the training GMMs, but unavailable in the training time. 
(d)-(e): The Langevin sampler with step size trained by amortized SVGD 
obtains close approximation with $T=5$ or $15$ steps.  
 (b)-(c): The typical power decay step size requires more Langevin iterations (and hence computation cost) to converge.  
%We use equation \eqref{equ:aksd} to perform amortized KSD and find that amortized KSD (f) does not work as well as amortized SVGD. 
%Amortized SV
}
%generated from the same distribution but 
%test them a new GMM (te
%Approximate GMM with 10 mixture components from the family distribution $\Q_1$. Comparison of kernel density estimation (KDE) constructed using 5,000 samples from true posterior, Langevin dynamics (LD) and Langevin sampler trained by amortized SVGD or KSD, respectively.
%Red: true density; Green: Langevin dynamics that runs for 15 and 1000 iterations (which denote as Langevin (15) or Langevin (1,000)) respectively; Blue: KDE with the $T = 5, 15$ Langevin sampler ( $T$ is the number of layers in equation \eqref{equ:ls}) trained by amortized SVGD; Grey: KDE with the $T=15$ layers of Langevin sampler trained by amortized KSD. }
\label{fig:gmm}
\end{figure*}

\begin{figure*}[htbp]
\centering
\begin{tabular}{cccc}
\raisebox{3.2em}{\rotatebox{90}{\small Log10 MSE}}\includegraphics[width=0.185\textwidth]{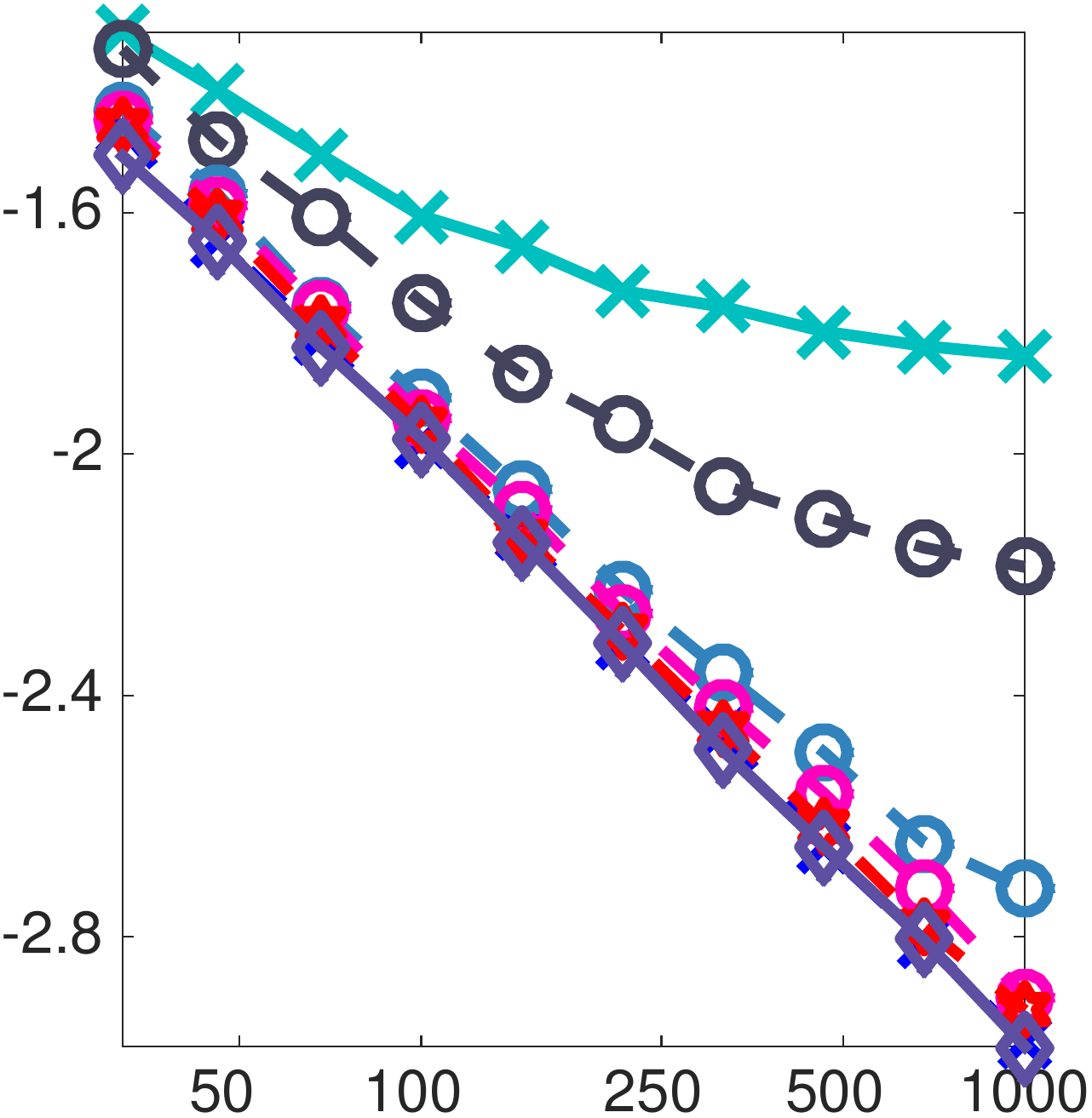} &
\includegraphics[width=0.185\textwidth]{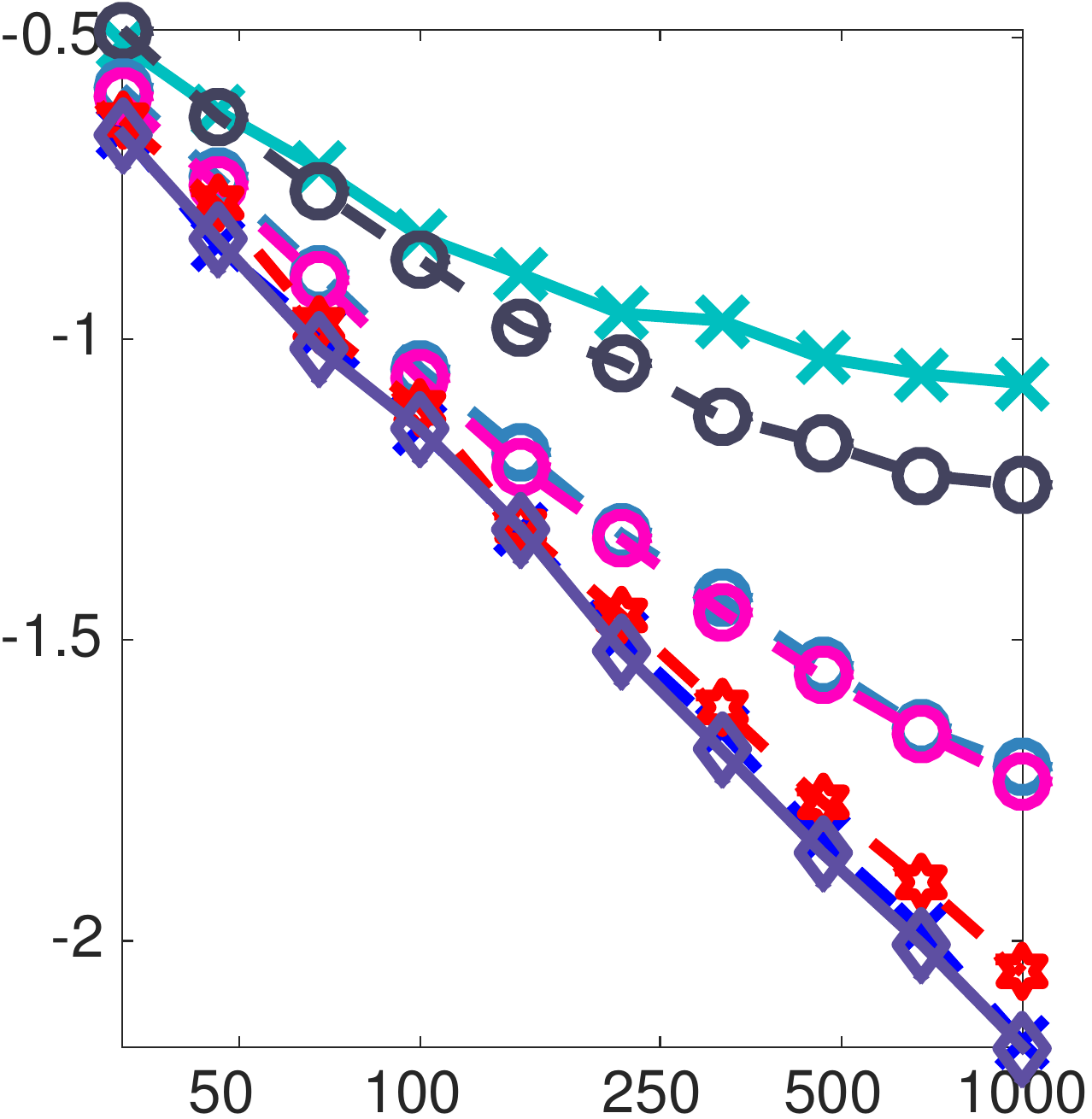} &
\includegraphics[width=0.185\textwidth]{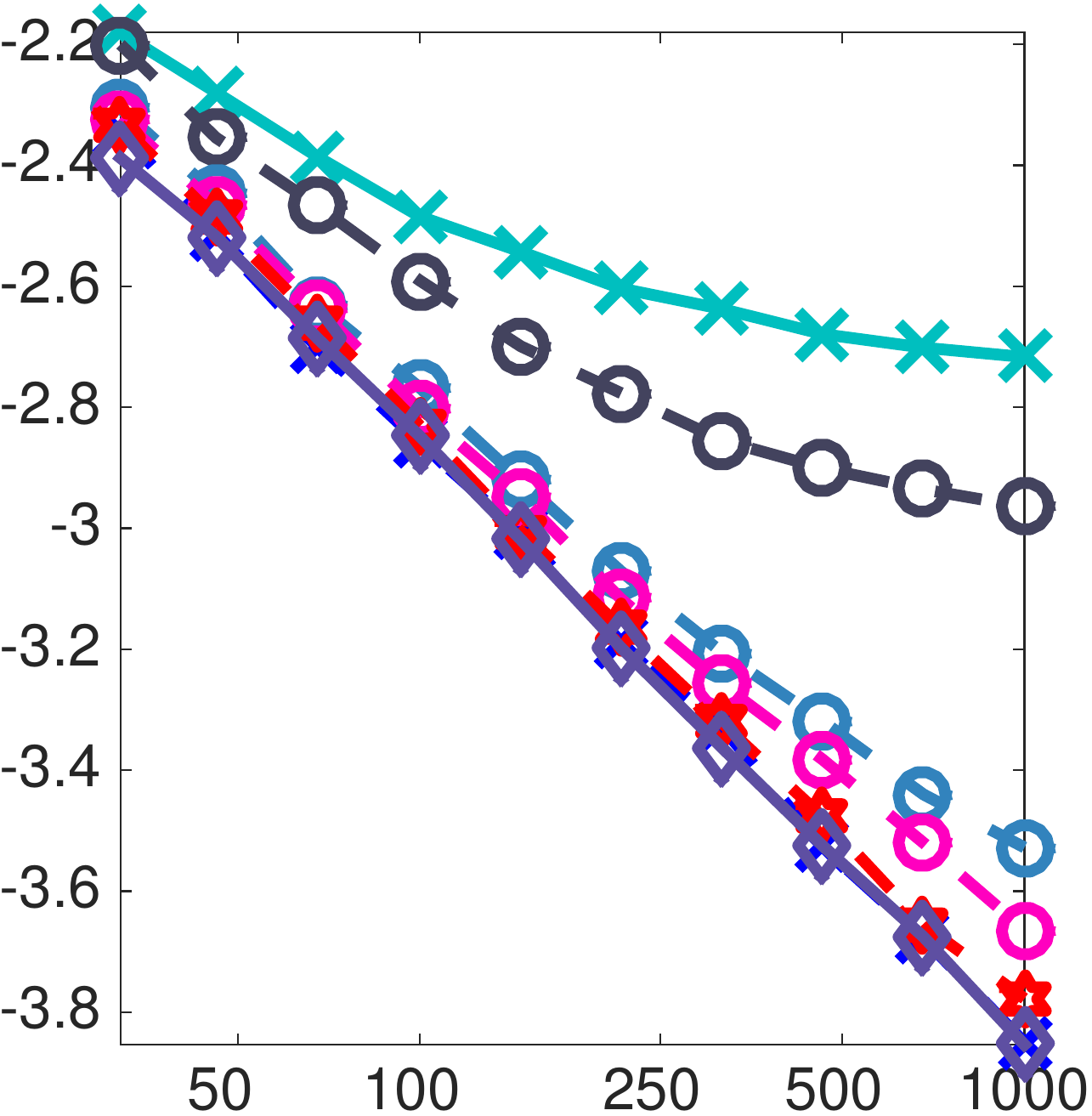} &
\raisebox{1em}{\includegraphics[width=0.30\textwidth]{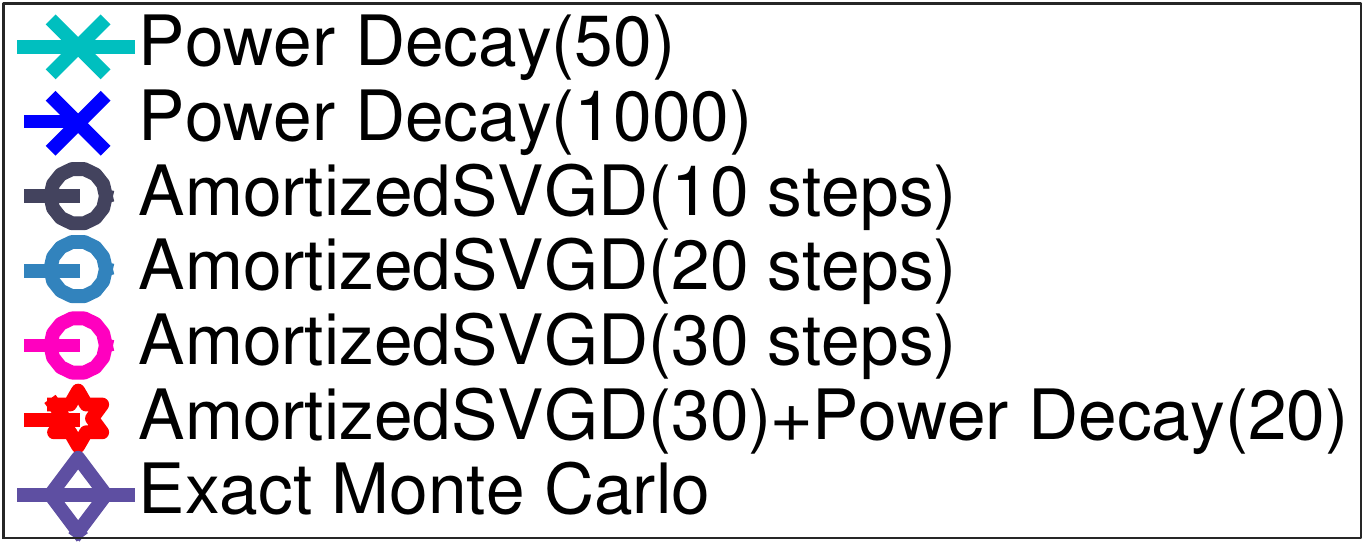}} \\
{\small Sample Size ($n$)} & {\small Sample Size($n$)} & {\small Sample Size ($n$)} & \vspace{.5em} \\
(a) $\E(x)$ & (b) $\E(x^2)$ & (c)  $\E(\cos(wx+b))$ &  \\
\end{tabular}
\vspace{-8bp}\caption{\small
Result on Gaussian-Bernoulli RBM. 
(a)-(c) show the mean square errors for estimating expectation $\E_p(h(x))$ using Langevin samplers with different numbers $T$ of steps, 
with either power decay step sizes or adaptive step size trained by amortized SVD. 
% Langevin ($k$) denotes running $k$ iterations of Langevin dynamics, and Layer ($k$) denotes $T=k$ 
%%layer Langevin sampler.
We take $h(x) = x^j, (x^j)^2$ and $\cos(wx^j+b)$ where $x^j$ denotes the $j$-th coordinate of vector $x$ and $w\sim \N(0,1)$ and $b\sim \mathrm{Uniform}(0, 2\pi)$ and report the average MSE 
across all dimensions $j=1,\ldots, d$, over 20 random trials.  
%Evaluating 
}
%(a)-(c) show the mean square errors estimating expectation $\E_p(h(x))$ for $h(x) = x, x^2$ and $\cos(wx+b)$ where $w\sim \N(0,1)$ and $b\sim \mathrm{Uniform}(0, 2\pi)$ and report the average MSE over 20 random trials.  Langevin ($k$) denotes running $k$ iterations of Langevin dynamics, and Layer ($k$) denotes $T=k$ layer Langevin sampler.}
\label{fig:rbm}
\end{figure*}

\section{EXPERIMENTS}

We apply amortized SVGD to the two different applications mentioned in section \ref{sec:apps}, 
%We show that our method allows us to efficiently amortize Bayesian computation and 
 and demonstrate that our method can train expressive inference networks to draw samples from intractable posterior distributions. For all our experiments, we use the standard RBF kernel $k(z, z')=\exp(- \frac{1}{h}|| z- z'||^2_2)$, and take the bandwidth to be $h = \mathrm{med}^2 / \log n$, where $\mathrm{med}$ is the median of the pairwise distance between the current points $\{z_i\}^n_{i=1}$. 
 We use update \eqref{equ:follow3} in Algorithm 1, which solves Eq \eqref{equ:follow1}-Eq \eqref{equ:follow2} using a single gradient step. 
 We find that using more gradient steps does not change the final results  significantly, 
 but may potentially increase the convergence speed (see Appendix \ref{sec:diffsteps}).

\subsection{Training Langevin Samplers}\label{sec:sampler}
In this section, 
we use amortized SVGD (Algorithm~\ref{alg:alg3}) to learn the step size parameters in the Langevin sampler in \eqref{equ:ls}.  
We test a number of distribution families, 
including Gaussian Mixture, Gaussian Bernoulli RBM, Bayesian logistic regression and Bayesian neural networks. 
In all the cases, we train the sampler 
with a set of ``training distributions'' and evaluate the sampler 
on ``test distributions'' that are not seen by the algorithm during training. 
We compare our method to the 
typical Langevin sampler 
with power decay step size, selected to be the best from $\eta^t =  10^a / (t + b)^{\gamma}$
where $\gamma = 0.55$, $a \in \{-6,...,2\}$, $b \in \{0,...,9\}$. 
% where 
%$a$ and $b$ and $\gamma$ is selected in the range of XXXX. 
%we show that with amortized SVGD and our proposed Langevin sampler, we are able to draw samples from a family distribution $\Q$. In the following experiments, we specify the $\Q$ to be Gaussian Mixture, Gaussian Bernoulli Restricted Boltzmann Machine and parameter distributions of Bayesian classifiers.

\paragraph{Gaussian Mixture}\label{para:gmm}
We first train the Langevin samplers 
to learn to sample from simple Gaussian mixtures. 
We consider a family of Gaussian Mixtures  
$q_\vartheta(z) = \frac{1}{10} \sum_{i=1}^{10} \N(z; \vartheta_i, 0.1^2)$, 
where $\vartheta$ is the mean parameter. 
%and the evaluation is done by 

%We used amortized SVGD or KSD to train the Langevin sampler using algorithm \ref{alg:alg3}. For each iteration, we draw a mini-batch of $\mu_i \sim \mathrm{Uniform}(-3, 3)$ and select a $p(\x;~\mu) \in \Q_1$ as input to update the sampler parameter. %After training, we expect the trained samplers can draw samples 
%we evaluate the Langevin sampler using new distribution $p(x ; ~\mu)$ with new values of $\mu_i \sim \mathrm{Uniform}(-3, 3)$ never seen by the training algorithm. 
%To evaluate performance of the Langevin sampler, we random select $p(x; ~\mu) \in Q_1$ that has not been used by the training process for test.
%\red{which mu is used in the graph?}
We train the sampler by drawing random elements of $\vartheta$ from $\mathrm{Uniform}(-1,1)$ at each iteration of Algorithm~\ref{alg:alg3},
and evaluate the quality of the samplers on new values of 
$\vartheta$ drawn from the same distribution, but not seen during training. 
%During training, we randomly draw elements of $\vartheta$ from $\mathrm{Uniform}(-1,1)$ at each iteration. 
%For evaluation, we show that the result of trained samplers in Figure~\ref{fig:gmm} by
%susing a new value of $\vartheta$ drawn from the same distribution, but not seen in the training time. 
%In Figure \ref{fig:gmm} we visualize the result of trained samplers using a new value of $\mu_i \sim \mathrm{Uniform}(-1,1)$ never seen by the training algorithm. 
In Figure~\ref{fig:gmm}, %visualizes the samples given by the trained samplers 
%on a typical $p_\vartheta$.  
we find that the sampler 
trained by amortized SVGD obtains good approximation 
with $T=5$ or $15$ steps of Langevin updates (Figure~\ref{fig:gmm}(d)-(e)). 
% amortized SVGD closely approximates $p_\vartheta$ with a $T=15$ layer Langevin sampler (Figure \ref{fig:gmm} (e)), 
In comparison, the typical Langevin sampler with the best power decay step size performs much worse (Figure~\ref{fig:gmm}(b)).

\begin{figure*}[ht]
\centering
\hspace{-.1\textwidth}\begin{tabular}{ccc}
\raisebox{3em}{\rotatebox{90}{\small Test Accuracy}}  \includegraphics[width=0.20\textwidth]{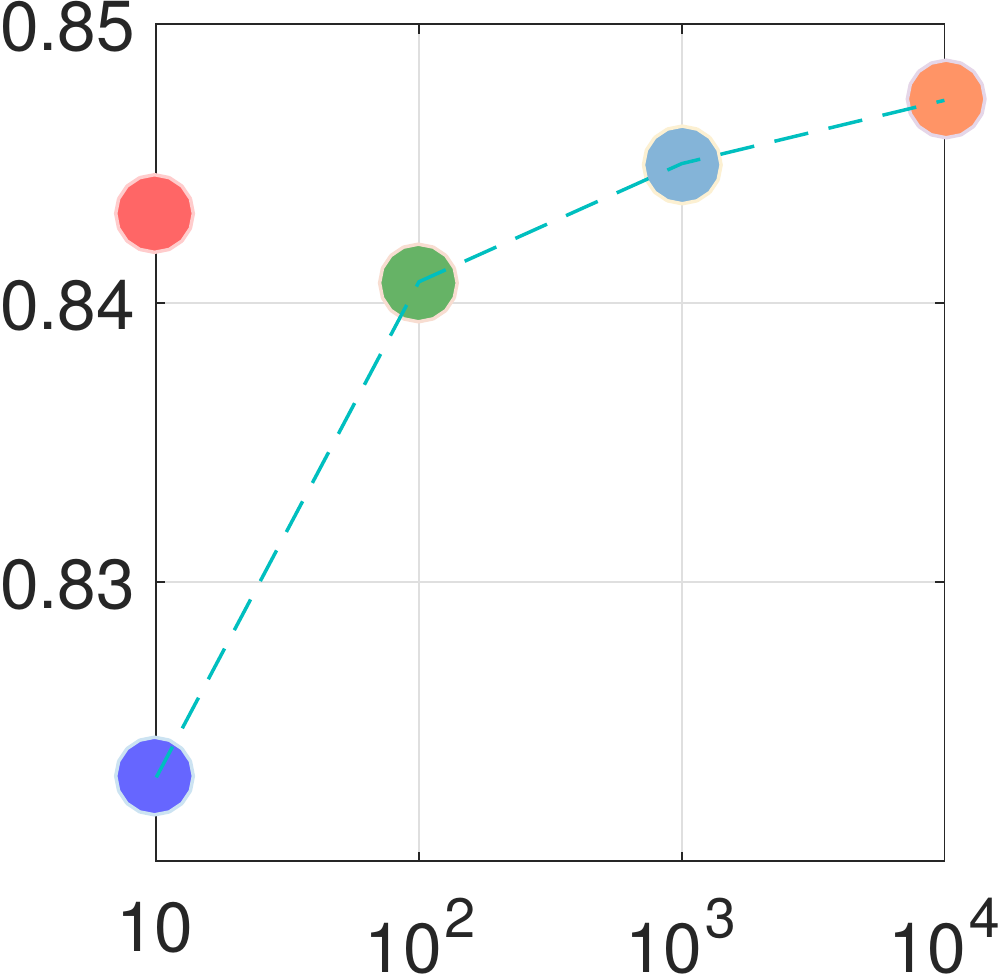}&
\raisebox{3em}{\rotatebox{90}{\small Test Accuracy}} \includegraphics[width=0.20\textwidth]{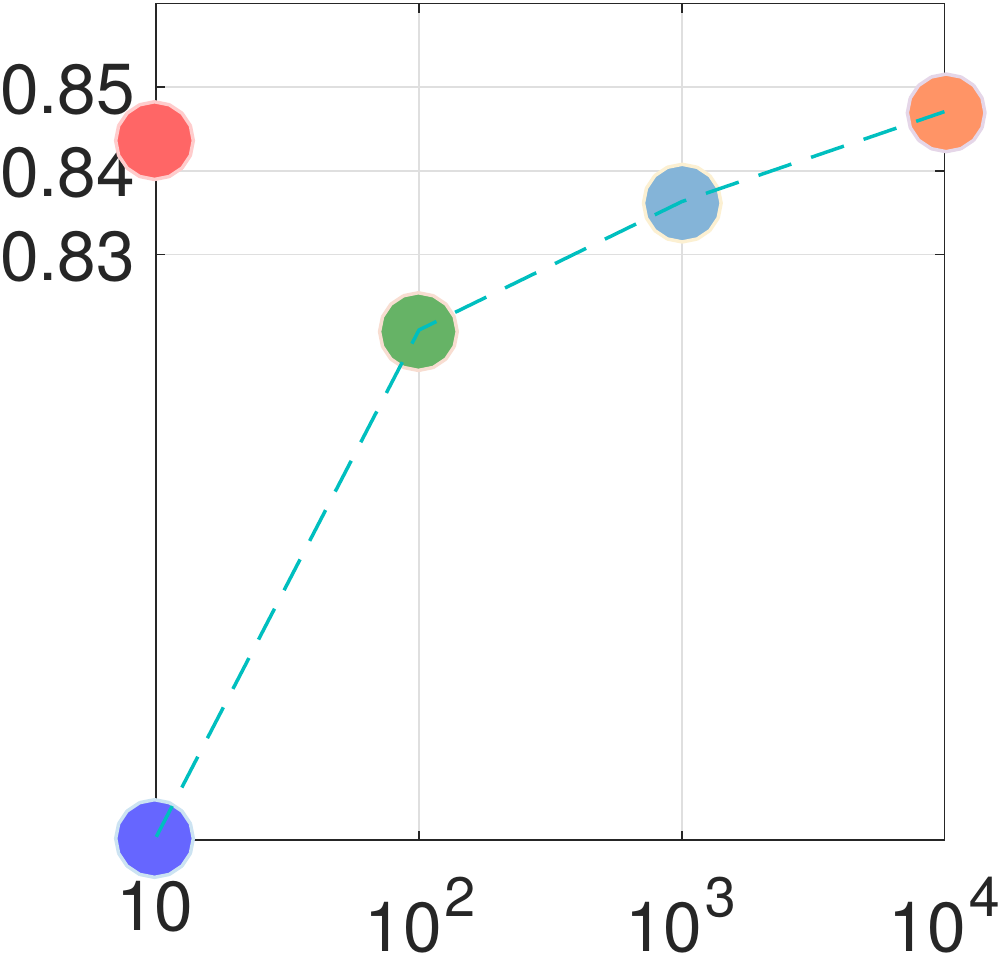} &
%\raisebox{8.3em}{\includegraphics[width=0.12\textwidth]{../figures/lr/lr_legend.pdf}} 
\raisebox{1.5em}{\includegraphics[width=0.016\textwidth]{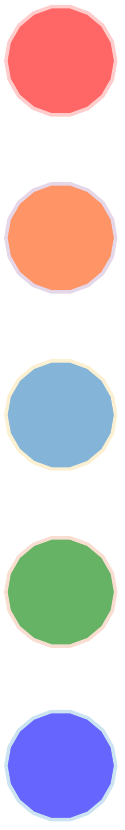} }
%\put(2, 71){\small Amortized SVGD (10 steps)}
\put(2, 60){\small Amortized SVGD (10 steps)}
\put(2, 49){\small Power Decay ($10^4$ steps)}
\put(2, 39){\small Power Decay ($10^3$ steps)}
\put(2, 28){\small Power Decay ($10^2$ steps)}
\put(2, 17){\small Power Decay (10 steps)} \\
\vspace{-.5\baselineskip}
\small ~~~~~~Steps of Langevin updates & \small ~~~~~~Steps of Langevin updates & \\
(a) \small Bayesian logistic regression & (b) \small Bayesian neural networks &
\end{tabular}
\vspace{-10bp}\caption{\small
By training the step size using amortized SVGD on dataset a9a, 
we obtain good performance ``on'' datasets a1a-a8a using Langevin sampler with only $T=10$ steps of Langevin updates. 
}
%Bayesian classification results on the adult datasets. LR-Layer(10) denotes Bayesian Logistic Regression parameters generated by a $T=10$ layer Langevin sampler, and NN-Layer(10) denotes Bayesian  Neural Network parameters generated by a $T=10$ layer Langevin sampler ,and Langevin($k$) denotes runing SGLD for $k$ iterations using best power decay step size.}
\label{fig:classification}
\end{figure*}

\paragraph{Restricted Boltzmann Machine (RBM)}\label{sec:rbm}
We test our method on 
Gaussian-Bernoulli RBM which
is high dimensional  and multi-modal. 
%is multi-modal and have highf%high dimensions and more complex multi-modal posteriors by choosing $\Q$ to be a family of Gaussian-Bernoulli Restricted Boltzmann Machine (RBMs).
Gaussian-Bernoulli RBM is a hidden variable model consisting of a continuous observable variable $z\in \R^d$ and a 
binary hidden variable $h \in \{\pm 1\}^{\ell}$ with joint probability
$p_\vartheta (z, h) \propto \exp( z^\top B h + b^\top z + c^\top h - \frac{1}{2} || z ||_2^2),$
where the parameters include $\vartheta = \{B, b, c\}.$ 
%where $Z$ is the normalization constant.
We obtain the marginal distribution of $p_\vartheta(z)$ by summing over $h$: %we can show that $P_\mathrm{rbm}(x)$ has the form
$p_{\vartheta}(z) \propto \exp\big [ b^\top z - \frac{1}{2}|| z ||_2^2 + \sigma(B^\top z + c) \big ],$
where $\sigma(h) = \sum_{i=1}^\ell\log (\exp(h_i) +\exp(- h_i))$. 
%\todo{the formula of score function}
%Following a similar setting as Gaussian mixture, we define the RBM family distribution as 
%$$
%\Q_2= \left\{P_{\mathrm{rbm}}(x; ~B, b, c) \mid B \sim \{\pm 0.1\}, ~b~\text{and} ~c \sim \N(0, \mathrm{I}^2)\right\}.
%$$
In our experiments, we take $\ell=10$ hidden variables 
and $d=100$ observable variables,
and randomly draw $b$ and $c$ from $\normal(0, I)$ and 
$B$ uniformly from $\{\pm 0.1\}$ in the training time. 
The evaluation is on a new set of parameters $\vartheta = \{B, b, c\}$ drawn from  the same distribution. 

Figure~\ref{fig:rbm} shows the result when we use the trained samplers to estimate integral quantities of form 
$\E_{p_\vartheta} [ h] $ with different testing functions $h$.  
The plots show the MSE for estimating $\E_{p_\vartheta} [h]$
using $\sum_i h(z_i)/n$ with $z_i$ generated from the trained sampler $z_i = f(\xi_i, p_\vartheta; ~\eta)$. 
%Note that once we finished training the sampler, we can generate as many sample as we want. This is 
Here the sample size $n$ is the number of i.i.d. samples generated 
from the trained sampler in the testing time. 
In the case of typical, non-adaptive Langevin samplers, it is the number of 
Langevin Markov chains that we run in parallel. 
%Here the 
In Figure \ref{fig:rbm} (a)-(c), 
we generally find that amortized SVGD allows us to train high quality Langevin samplers with a small number $T$ of Langevin update steps. 
We observe that we can always further refine the 
result of our trained samplers using additional typical MCMC steps. 
For example, in Figure  \ref{fig:rbm}, 
we find that when using 20 steps of typical Langevin dynamics (with a power decay step size)
 to refine the output of the $T=30$ layer Langevin sampler trained by amortized SVGD achieves results close to the exact Monte Carlo. 
% , we can achieve similar performance as converged SGLD (best power decay step size), the distribution of which is close to the true distribution (MC).

\paragraph{Bayesian Classification}\label{sec:bayes}
We test our method on Bayesian Logistic Regression and Bayesian neural networks for binary classification on real world datasets. 
In this case, the distribution of interest 
has a form of $p_\vartheta(z) = p(z | D)$, 
where $z$ is the network weights in logistic regression and neural networks, 
and $D$ is the dataset for binary classification, which we view as the parameter $\vartheta$, 
that is, different dataset $D$ yields different posterior $p(z |D)$, 
and we are interested in training the Langevin sampler
on a set of available datasets, 
and hope it performs well on future datasets that have similar structures. 
This setting can be useful, for example, 
in the streaming setting where we use 
existing datasets to adaptively improve the Langevin sampler. 
In our experiment, 
we take nine similar datasets (a1a-a9) from the libsvm repository\footnote{\url{https://www.csie.ntu.edu.tw/~cjlin/libsvmtools/datasets/}}; 
we train our Langevin sampler on a9a, and evaluate the sampler on the remaining 
8 datasets (a1a-a8a). Our training and evaluation steps are as follows: 

1. Estimate the step sizes $\{\eta^t\}_{t=0}^{T-1}$ of the Langevin sampler using amortized SVGD based on dataset a9a. 

2. Apply the Langevin sampler with the estimated step size to the training subsets of a$k$a, $k=1,\ldots,8$, to obtain posterior samples $\{z_i^k\}$ of the classification weights. 

3. Calculate the test likelihood of $\{z_i^k\}$ on the testing subsets of  a$k$a, $k=1,\ldots,8$. Report the averaged testing likelihood averaged on the 8 datasets in Figure \ref{fig:classification}. 

%Figure \ref{fig:classification} shows the average test log likelihood dataset 
% a1a-a8a with $T =10$ Langevin samplers obtained by amortized SVGD on a9a (we applied the Langevin sampler on the training subset of aka, $k=1,\ldots,8$ to get )
Because each dataset is relatively large, we use the stochastic gradient approximation as suggested by \citet{welling2011bayesian} (with a minibatch size of 100) in Langevin samplers.  
We find that the $T = 10$ Langevin samplers trained by amortized SVGD on a9a
is comparable with the $T = 10^3$ Langevin sampler with the best power decay step size. 

\subsection{Training VAE With Amortized SVGD}\label{steinvae}
We compare the entropy regularized VAE trained with amortized SVGD (denoted by \textit{ESteinVAE}), 
which is Algorithm~\ref{alg:alg2} with update \eqref{equ:update11alpha}, 
with the standard VAE and entropy regularized standard VAE (denoted by \textit{EVAE}) on the dynamically binarized MNIST dataset \citep{burda2015importance}. 
We tested the following settings: % includes: 

1. A standard VAE (VAE-f) with a fully connected encoder consisting of one hidden layer with 400 hidden units, and a Gaussian output hidden variable with diagonal covariance. 

2. A standard  convolutional Gaussian VAE (VAE-CNN) with a convolutional encoder consisting of 2 convolution layers with $5\times 5$ filters, stride 2 and $[16, 32]$ features maps,
followed by a fully connected layer with 512 hidden units. 

3. A convolutional Gaussian entropy regularized (EVAE-CNN) with the same encoder and decoder structures as the standard convolutional VAE (VAE-CNN).

4. Entropy regularized VAEs trained by our amortized SVGD in Algorithm~\ref{alg:alg2}, 
% (denoted by , respectively), 
 with the same encoder architecture as VAE-f and VAE-CNN, respectively, but removing the Gaussian noise on the top layer and adding a multiplicative Bernoulli noise with a dropout rate of 0.3 to the each layer of the encoder. These two cases are denoted by ESteinVAE-f, and ESteinVAE-CNN, respectively. 
%For benchmark both fully connected (VAE-f) and convolutional Gaussian VAEs (VAE-CNN) with $\mathrm{dim}(\z) = 32$ latent variable are tested, where the encoder and decoder has symmetric architecture: the fully connected encoder has one hidden layer with 400 hidden units, and a Gaussian output hidden variable with diagonal covariance; 
%the convolutional encoder consists of 2 convolution layers with $5\times 5$ filters, stride 2 and $[16, 32]$ features maps, followed by a fully connected layer with 512 hidden units. 
%As for the encoder of SteinVAE, we consider a non-Gaussian encoder which simply applies multiplicative Bernoulli noise with dropout rate 0.2 to the input layer, both for fully connected (SteinVAE-f) and convolutional structures (SteinVAE-CNN), and use same MLPs as benchmark VAEs respectively. 
%\begin{comment}
%Different from the Gaussian encoder, the latent variable of the non-Gaussian encoder is not parameterized with two outputs which represent mean and diagonal variance %respectively.
%\end{comment}

In all these cases, we use $\mathrm{dim}(\z) = 32$ latent variables and the decoders have symmetric architectures as the encoders in all models. Adam %\citep{kingma2014adam} 
is used with a learning rate tuned on the training images. 
For each training iteration, a batch of 128 images is used, 
and for each image we draw a batch of $m=5$ samples to apply amortized SVGD.

\paragraph{Marginal likelihood}
Table~\ref{tbl:mnist} reports the test log-likelihood of all the methods estimated using Hamiltonian annealed importance sampling (HAIS) \citep{wu2016quantitative} 
with 100 independent AIS chains and 10,000 intermediate transitions, averaged on 5000 test images. 
We find that ESteinVAE-f significantly outperforms VAE-f, 
and ESteinVAE-CNN slightly outperforms VAE-CNN and EVAE-CNN. 
Table~\ref{tbl:mnist} also reports the effective sample size (ESS) of the HAIS estimates. 
The fact that the effective sample sizes of all the methods are close suggests that accuracy of the different NLL estimates are comparable. 
%which are generally high and suggests that our likelihood estimation is reliable. 
%We can see that 
% to compute the test negative log-likelihood scores, which only needs the trained decoder networks for evaluation.
%Next we report the test log-likelihood results in Table \ref{tbl:mnist}.Unlike Gaussian encoder based VAEs, which can use importance sampling \citep{kingma2013auto} to estimate the marginal log-likelihood, the latent variables of more complex non-Gaussian inference networks do not have a assumed distribution such as multivariate Gaussian.We use Hamiltonian annealed importance sampling (HAIS) \citep{wu2016quantitative} to compute the test negative log-likelihood scores, which only needs the trained decoder networks for evaluation.
%We evaluate 5000 test images with 100 independent AIS chains, 10,000 intermediate transitions, with which the AIS chains usually converge. 
%Under the same experiment settings, for both fully connected and convolutional versions, SteinVAEs outperform benchmark VAEs. 
%To evaluate the reliability of the result evaluated by HAIS, we compute the effective sample size(ESS), where these values are very high in all these four models. What's more, the values of ESS acquired by AIS are higher than that uses importance samples drawn from Gaussian encoders \citep{li2017approximate}.
%Some random samples for mnist in Figure \ref{fig:vae_samples}. We see that our approachproduces samples that are visually similar to VAE samples, which demonstrate amortized SVGD can achieve state of the art results of training deep generative models.

\begin{table}[t]
\caption{\small Negative log-likelihood on binarized mnist test dataset.}
\centering
\vspace{3bp}\begin{tabular}{ccc}
\hline
\textbf{Model} & \textbf{NLL/nats} & \textbf{ESS} \\\hline
\small VAE-f & 90.32 &  84.11 \\%  90.10 
\small ESteinVAE-f & 88.85 & 83.49 \\ % 90.88
\hline
\small VAE-CNN & 84.68 & 85.50 \\ 
\small EVAE-CNN & 84.43 & 84.91 \\
\small ESteinVAE-CNN & 84.31 & 86.57 \\% 83.85 \\
\hline
\end{tabular}
\label{tbl:mnist}
\end{table}

\begin{table}[t]
\caption{\small Quantitative imputation experiment based on 2,000 true images and 500 reconstructed images per true image.} 
\centering
\vspace{3bp}\begin{tabular}{ccc}
\hline
\textbf{Model} & \textbf{Accuracy} & \textbf{Entropy} \\
\hline
\small ESteinVAE-CNN   & 0.84			   & 0.501 \\
%\hline
\small EVAE-CNN 		& 0.82			   & 0.382 \\
%\hline
\small VAE-CNN        & 0.83              & 0.340   \\
\hline
\end{tabular}
\label{tb:imputation}
\end{table}

\begin{comment}
\begin{table}[t]
\caption{\small Quantitative imputation experiment based on 2,000 true images and 500 reconstructed images per true image.} 
\centering
\vspace{3bp}\begin{tabular}{ccc}
\hline
\textbf{Model} & \textbf{Accuracy} & \textbf{Entropy} \\
\hline
\small ESteinVAE-CNN   & 0.84			   & 0.570 \\
%\hline
\small EVAE-CNN 	   & 0.83		   & 0.442 \\
%\hline
\small VAE-CNN        & 0.82	    & 0.348   \\
\hline
\end{tabular}
\label{tb:imputation}
\end{table}

If I increase the imputation area, we can see that both the entropy term of EVAE and SteinVAE increases but the entropy of VAE stays, which makes sense because we now have larger uncertainty. For Accuracy, there may be two reasons that cause mismatching (generate more diverse images from the imputated ambiguous images or all the generated images are the same label but different with original image label.) The first one may cause the low accuracy in SteinVAE. For VAE, there still may exsits some images, even with imputation, it still can have a good reconstructed images (for example digit "1"), so we can not only see the image accuracy. What's mroe. the imputation process did not suggest that we should have a higher accuracy(each generated images are indepdent). 

In the new experiment, the entropy of the VAE stays while the accuracy drops a lot, which indicates it generates images with the same label but different with original one.
\end{comment}

\begin{figure}[t]
\centering
\begin{tabular}{c}
\includegraphics[width=0.45\textwidth]{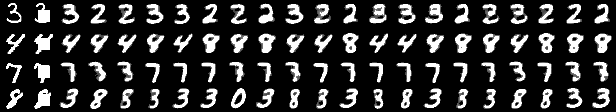} \\
\small ESteinVAE-CNN \\
\includegraphics[width=0.45\textwidth]{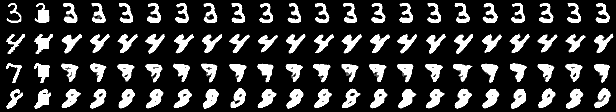} \\
\small EVAE-CNN \\
\includegraphics[width=0.45\textwidth]{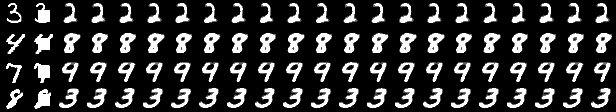} \\
\small VAE-CNN\\
\end{tabular}
\vspace{-10bp}\caption{\small Imputation results from benchmark VAE-CNN, EVAE and ESteinVAE-CNN. The first column shows the original image. Col. 2 shows the missing image.
The remaining columns show imputations for 20 reconstructed samples.}
\label{fig:denosing}
\end{figure}

\paragraph{Missing data imputation}
We demonstrate that ESteinVAE is able to learn multi-modal latent representation with the binary dropout noise.  
%We first test the ability to learn uncertainty using amortized SVGD. 
We consider missing data imputation with pixel missing in a square sub-regions of the image (the second column of Figure~\ref{fig:denosing}). 
We use a simple method to reconstruct image by applying one step of encode-decode operation starting from uniform noise $\mathrm{Uniform}(0, 1)$. 
%approximate Gibbs sampler to 
%we first fill the missing part of the image with $\mathrm{Uniform}(0, 1)$,  and then 
%where a square region of the image is missing. 
%First, we use a random noise drawn from $\mathrm{Uniform}(0, 1)$ to fill the missing square region. Then we do 1 step reconstruction from the missing data. Notice that our imputation procedure is different from that in \citet{rezende2014stochastic}. 
Specifically, let $x = [x^{v}, x^m]$ where $x^v$ and $x^m$ denotes the visible and missing parts, respectively. Our imputation procedure consists of
the following  steps: 
%with the whole input image denoted as $x_c$, our imputation procedure consists following two steps:
\begin{enumerate}

\item Draw $x^c$ from $\mathrm{Uniform}(0, 1)$;

\item Draw $z \sim q_\phi(z | x^c,~x^{v})$; 

\item Draw $x \sim p(x | z)$.
\end{enumerate}

Each column (starting from the third column) of Figure~\ref{fig:denosing} shows an independent run of this procedure, 
where we can see that ESteinVAE-CNN is able to generate diverse construction when ambiguity exists, 
while the EVAE-CNN and VAE-CNN tend to be trapped in a local mode. 
This suggests that the diagonal variance of the latent variables in the Gaussian encoder of VAE-CNN tends to be small, 
underestimating the posterior uncertain, 
while ESteinVAE can capture the multi-modal posterior due to the dropout noise.

Table \ref{tb:imputation} is the quantitative result of the imputation experiment, 
in which  the ``accuracy'' column denotes the number of original images whose digit is in its reconstructed images.
%the number of times the reconstructed images belonging to the same digit as the original image, 
and the ``entropy'' column denotes  the entropy of the probability of the reconstructed images belonging to different digit classes. 
EsteinVAE obtains more diverse images and slightly more accurate reconstructed images.

\section{CONCLUSION}

We propose a new method to train neural samplers for given distributions, together with various applications to learning to draw samples using neural samplers. Future directions include exploring more efficient neural architectures and theoretical understanding of our method.

{\bf Acknowledgments}\\ 
This work is supported in part by NSF CRII 1565796. We thank Yingzhen Li from University of Cambridge for her valuable comments and feedbacks.

\bibliographystyle{icml2017}
\bibliography{bibrkhs_stein.bib}

%!TEX root = ../main.tex

\clearpage 

\appendix

%!TEX root = ../main.tex

\section{KSD Variational Inference}
\label{sec:ksd}

\begin{figure}[t]
\centering
\begin{tabular}{ccc}
\includegraphics[width=0.14\textwidth]{figures/gmm/posterior/truth.pdf} &
\includegraphics[width=0.14\textwidth]{figures/gmm/posterior/layer15.pdf} &
\includegraphics[width=0.14\textwidth]{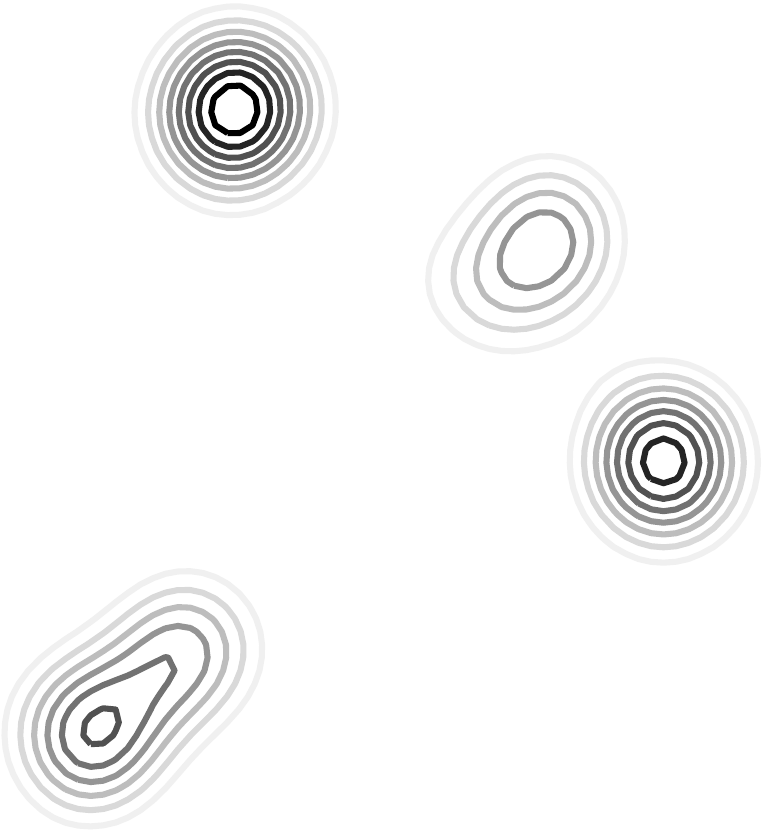}  \\
\fttmp \btt (a) Test Distribution \\(truth) \ett 
  & \fttmp \btt (b)  Amortized SVGD\\ (15 steps) \ett 
 & \fttmp \btt (c) Amortized KSD \\ (15 steps) \ett
\end{tabular}
\vspace{-10bp}\caption{\small 
Learning to sample from GMM.
The Langevin sampler 
with step size trained by amortized SVGD (b)
obtains close approximation with $T=15$ is close to the true test distribution (a) while
amortized KSD (c) which we use equation \eqref{equ:aksd} to perform does not work as well as amortized SVGD.
%while that the typical power decay step size reduces more iterations (and hence computation) to converge (b)-(c). 
%We use equation \eqref{equ:aksd} to perform amortized KSD and find that amortized KSD (f) does not work as well as amortized SVGD. 
%Amortized SV
}

\label{fig:ksd_compare}
\end{figure}

Kernelized Stein discrepancy (KSD) provides a discrepancy measure between distributions 
and can be in principle used as a variational objective function in replace of KL divergence. 
In fact, thanks to the special form of KSD (\eqref{equ:ksdexp}-\eqref{equ:uv}), 
one can derive a standard stochastic gradient descent for minimizing KSD without needing to estimate $q_\eta(z)$ explicitly, 
%the typical gradient-based optimization can be performed without needing to estimate $q_\eta(z)$ explicitly. 
which provides a conceptually simple wild variational inference algorithm. 
Although %we observe that 
this work mainly focuses on 
amortized SVGD which we find to be easier to implement and tend to perform superior to KSD variational inference in practice (see Figure~\ref{fig:ksd_compare}),   
we think the KSD approach is of theoretical interest and hence give a brief discussion here. %in this section. 
%here we include a brief discussion because its own 
%In practice, we did not observe this approach
%Here we give a brief discussion on this approach and connect it with amortized SVGD and operator variational inference \citep{XXX}. 

%and The simple form of KSD shown in \eqref{equ:} makes it possible 
%Amortized SVGD minimizes the $\mathrm{KL}$ divergence objective. 
%We provide an alternative amortized method based on directly minimizing the kernel Stein discrepancy (KSD) \citep{liu2016kernelized}. Thanks to the special form of KSD, the typical gradient-based optimization can be performed without needing to estimate $q(x)$ explicitly.
%Based on similar idea, we can perform amortized inference by minimizing KSD.
Specifically, take $q_\eta$ to be the density of the random output $\z = f(\xi;~\eta)$ when $\xi\sim q_0$, 
and we want to find $\eta$ to minimize $\S(q_\eta~||~p)$. 
Assuming $\{\xi_i\}$ is i.i.d. drawn from $q_0$, we can approximate $\S^2(q_\eta~||~p)$ unbiasedly 
using the U-statistics in \eqref{equ:uv}, and derive a standard gradient descent
%\begin{align*}
% \S^2(q_\eta~||~ p) \approx \frac{1}{n(n-1)}\sum_{i\neq j} \kappa_p(f(\xi_i;~\eta), ~ f(\xi_j;~\eta)), 
%\end{align*}
%for which a standard gradient descent can be derived for optimizing $\eta$:
\begin{gather}\label{equ:aksd}
\eta \gets \eta -   \epsilon \frac{2}{n(n-1)}\sum_{i\neq j} \partial_\eta f(\xi_i;~\eta) ~ \nabla_{\z_i}\kappa_p(\z_i, \z_j),
%\eta \gets \eta +  \epsilon \frac{2}{n(n-1)}\sum_{i\neq j} \partial_\eta f(\eta;~\xi_i) \nabla_\z\kappa_p(f(\eta;~\xi_i), ~ f(\eta;~\xi_j)).
%\eta \gets \eta +  \epsilon  \nabla_\eta \hat \S^2(q_\eta~||~ p), \\
%\nabla_\eta \hat \S(q_\eta~||~ p) = \frac{2}{n(n-1)}\sum_{i\neq j} \partial_\eta f(\eta;~\xi_i) \nabla_\z\kappa_p(f(\eta;~\xi_i), ~ f(\eta;~\xi_j)).
\end{gather}
where $\z_i = f(\xi_i;~\eta)$.
%where 
%whose derivative w.r.t. $\eta$ is
%$$
%\nabla_\eta \hat \S(q_\eta~||~ p) = \frac{2}{n(n-1)}\sum_{i\neq j} \partial_\eta f(\eta;~\xi_i) \nabla_\z\kappa_p(f(\eta;~\xi_i), ~ f(\eta;~\xi_j)). 
%\nabla_\eta \hat \S(q_\eta~||~ p) = \frac{2}{n(n-1)}\sum_{i\neq j} \partial_\eta f(\eta;~\xi_i) \nabla_\z\kappa_p(f(\eta;~\xi_i), ~ f(\eta;~\xi_j)). 
%$$
This enables a wild variational inference method based on directly minimizing $\eta$ with standard (stochastic) gradient descent. 
We call this algorithm \emph{amortized KSD}.  
%See Algorithm~\ref{alg:alg1}. 
Note that \eqref{equ:aksd} is similar to \eqref{equ:follow3} in form, 
but replaces $\ff^*(\z_i)$ with 
$$\bar{\ff}^*(\z_i)  \overset{def}{=}  - 2\sum_{j\colon i\neq j} \nabla_{\z_i}\kappa_p(\z_i, \z_j)/(n(n-1)).$$   
Here $\bar \ff^*$ depends on the second order derivative of $\log p$ because $\kappa_p(z,z')$ depends on $\nabla \log p$, 
which makes it more difficult to implement amortized KSD than amortized SVGD. 
%Here it is more 
%We call this algorithm amortized KSD. 

Intuitively, 
minimizing KSD can be viewed as 
seeking a stationary point of KL divergence under SVGD updates. 
%minimizing the amount of decrease of KL divergence that SVGD update can yield. 
%minimizing a contrastive divergence objective function. 
To see this, recall that $q_{[\epsilon \ff]}$ denotes the density of $\z' = \z+\epsilon \ff(\z)$ when $\z  \sim q$. 
From \eqref{equ:ff00}, we have for small $\epsilon$, 
%Combining \eqref{equ:ff00} and \eqref{equ:ksdexp}, we can show that 
 %RThen we have 
\begin{align*} 
\S^2(q ~||~ p) 
 \approx \frac{1}{\epsilon}\max_{\phi\in \F}\big\{{\KL(q ~|| ~p) - \KL(q_{[\epsilon\ff]} ~||~ p)}\big\}.  
\end{align*}
That is, KSD measures the maximum degree of decrease in the KL divergence when we update the particles along the optimal SVGD perturbation direction $\ff^*$. % given by \eqref{equ:ff00}. 
If $q = p$, then the decrease of the KL divergence equals zero and $\S^2(q~||~p)$ equals zero. 
In fact, 
%as shown in, 
KSD can be explicitly represented as the magnitude of the functional gradient of the KL divergence w.r.t. $\ff$ in RKHS  \citep{liu2016stein}, 
 $$
\S(q~||~p) = \Big|\Big| 
\nabla_\ff  F(0) %~ \big |_{\ff = 0} 
%\frac{d}{d \ff} F(\ff)\big |_{\ff = 0} 
\Big|\Big|_{\H^d}, ~~~~~ F(\ff) \overset{def}{=}  \KL(q_{[\epsilon\ff]} ~|| ~ p) , 
$$
%where $q_{[\ff]}$ is the density of $\z = \z + \ff(\z)$ when $\z\sim q$, and 
where $ \nabla_{\ff}F(\ff)$ denotes the functional gradient of the function $F(\ff)$ w.r.t. $\ff$ defined in RKHS $\H^d$, and $ \nabla_{\ff}F(\ff)$ is also an element in $\H^d$. 
Therefore,   
in contrast to amortized SVGD which attends to minimize the KL objective $F(\ff)$, 
KSD variational inference minimizes the gradient magnitude $||\nabla_\ff  F(0)  ||_{\H^d}$ of KL divergence. 
%can be treated as explicitly minimizing the magnitude of the gradient of the KL divergence, 
%in contrast to amortized SVGD which attends to minimize the KL divergence objective itself. 
%In connection with 
% KSD variational inference can be treated as minimizing the magnitude of the gradient of KL divergence.

%It is interesting to provide new method for wild variational inference. However, in practice amortized KSD turns to be not as good as amortized SVGD. Detail experiment and analysis can be found in section \ref{sec:sampler}.

This idea is closely related to 
the operator variational inference \citep{operator}, which directly minimizes the variational form of Stein discrepancy in \eqref{equ:ff00} and \eqref{equ:ksd} 
with $\F$ replaced by sets of parametric neural networks. 
%on parametric f
Specifically, 
\citet{operator} assumes $\F$ consists of a neural network $\ff_\tau(\z)$ with parameter $\tau$, and find $\tau$ jointly with $\eta$ by solving 
a min-max game: 
$$\min_{\eta} \max_{\tau} \E_{\z\sim q_\eta} [\sumsteinp \ff_\tau(\z)].$$ 
This yields a more challenging computation problem, although it is possible that the neural networks provide stronger discrimination than RKHS in practice.  
%In contrast, %by leveraging the closed form solution in RKHS, 
The main advantage of the KSD based approach is that it leverages the closed form solution in RKHS, 
yields a simpler optimization formulation based on standard gradient descent.  
%instead of a min-max problem that can be computationally more expensive, or have difficulty in achieving convergence. 
%that cause additional computational cost a
%avoids the minimization of an additional parameter $\tau$. 
%Because $\kappa_p(x,x')$ (defined in \eqref{equ:kp}) depends on the derivative  $\nabla_x \log p(x)$ of the target distribution, the gradient in \eqref{equ:aksd} depends on the Hessian matrix $\nabla_x^2 \log p(x)$ and is hence less convenient to implement compared with amortized SVGD (the method by \citet{operator} also has the same problem). However, this problem can be alleviated using automatic differentiation tools, which be used to directly take the derivative of the objective in \eqref{equ:ueta} without manually deriving its derivatives. 

Figure \ref{fig:ksd_compare} shows results of Langevin samplers trained by amortized SVGD and amortized KSD, respectively,  
for learning simple Gaussian mixtures under the same setting as that in Section \ref{sec:sampler}. 
We find that amortized KSD tends to perform worse (Figure~\ref{fig:ksd_compare}(c)) than amortized SVGD (Figure~\ref{fig:ksd_compare}(b)); 
given that it is also less straightforward to implement amortized KSD (for requiring calculating $\nabla_{z}\kappa_p(z,z')$ in \eqref{equ:aksd}), we did not test it in our other experiments.

\begin{figure*}[htbp]
\centering
\begin{tabular}{cccc}
\raisebox{2.em}{\rotatebox{90}{\small Log10 MSE}}\includegraphics[width=0.255\textwidth]{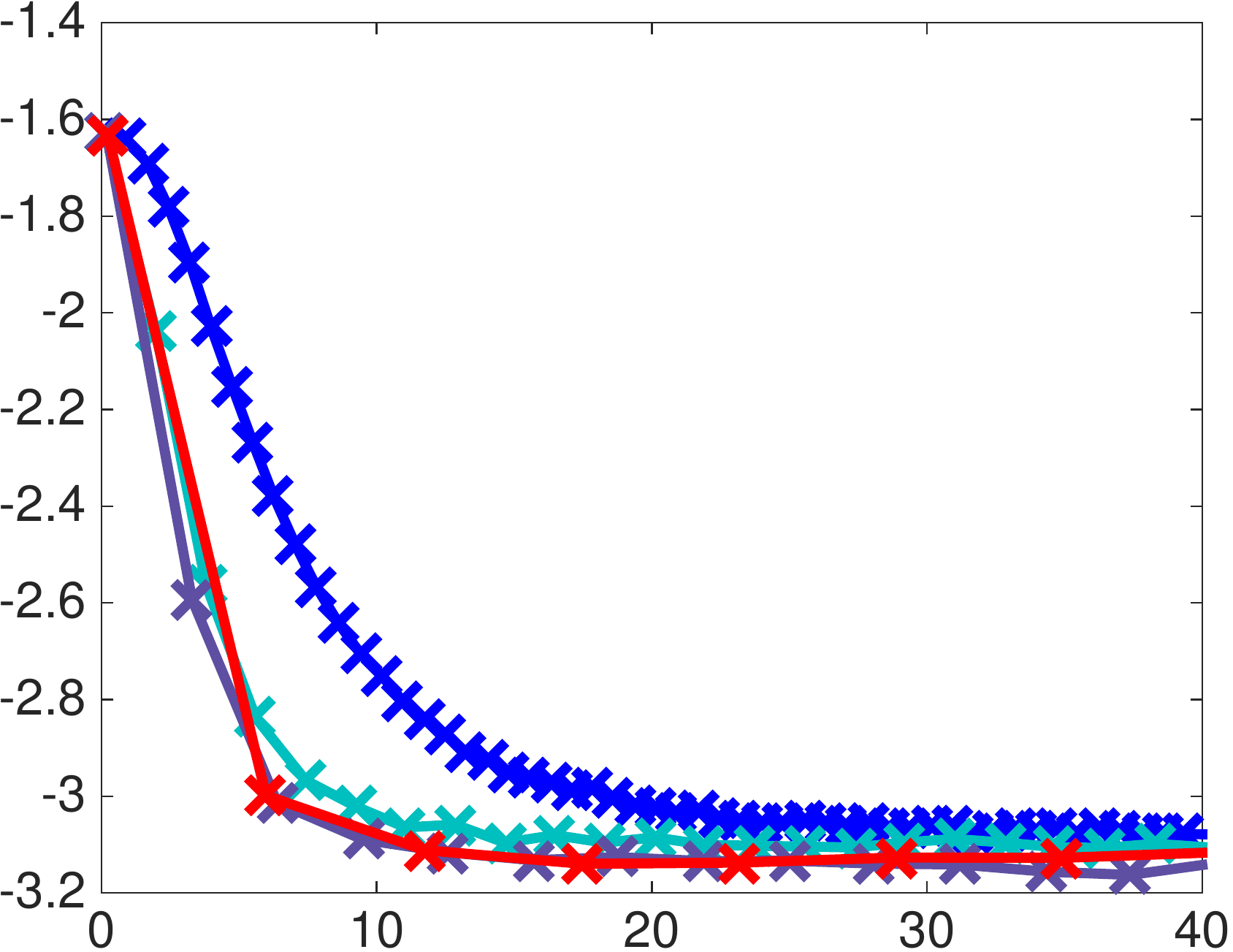} &
\includegraphics[width=0.255\textwidth]{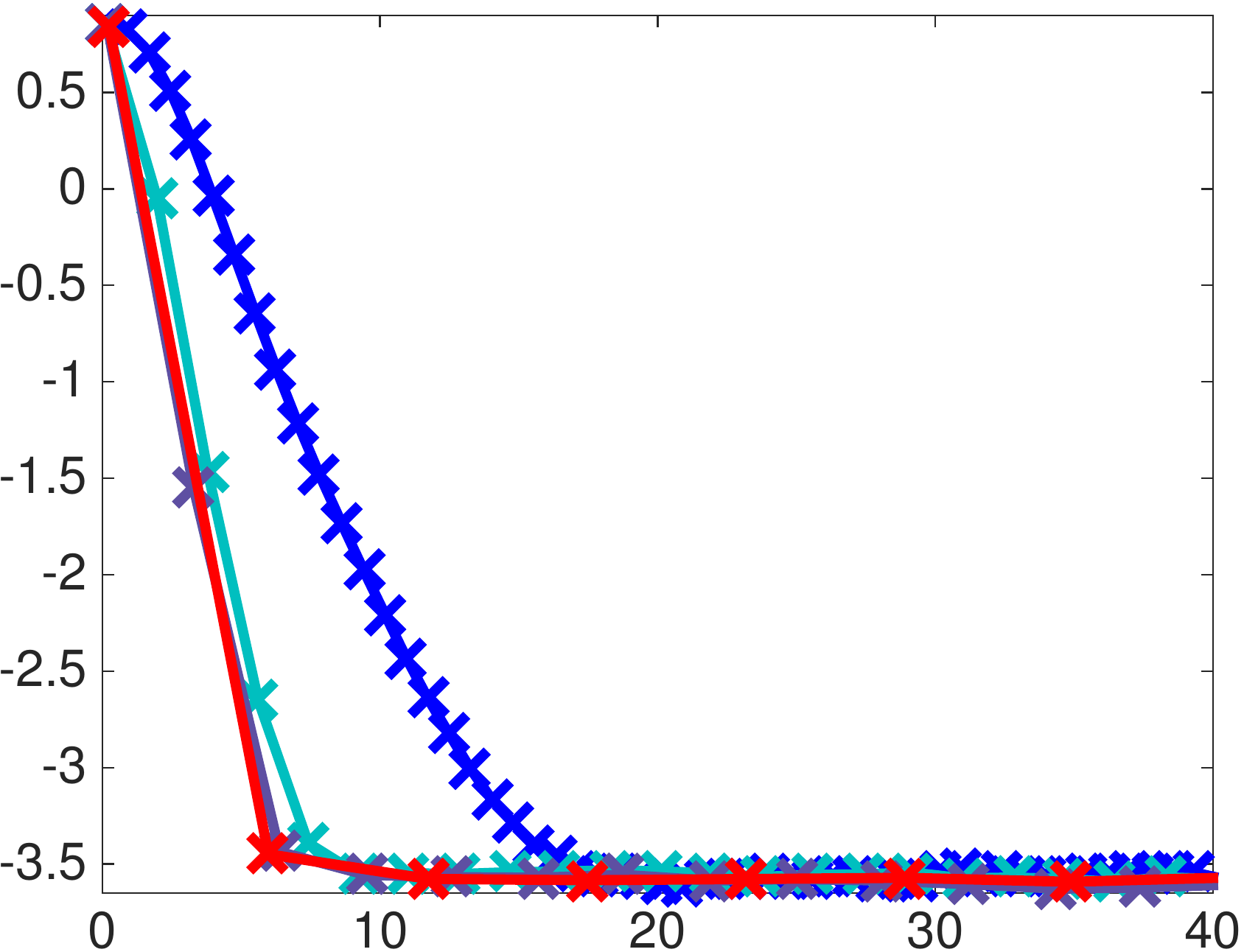} &
\includegraphics[width=0.255\textwidth]{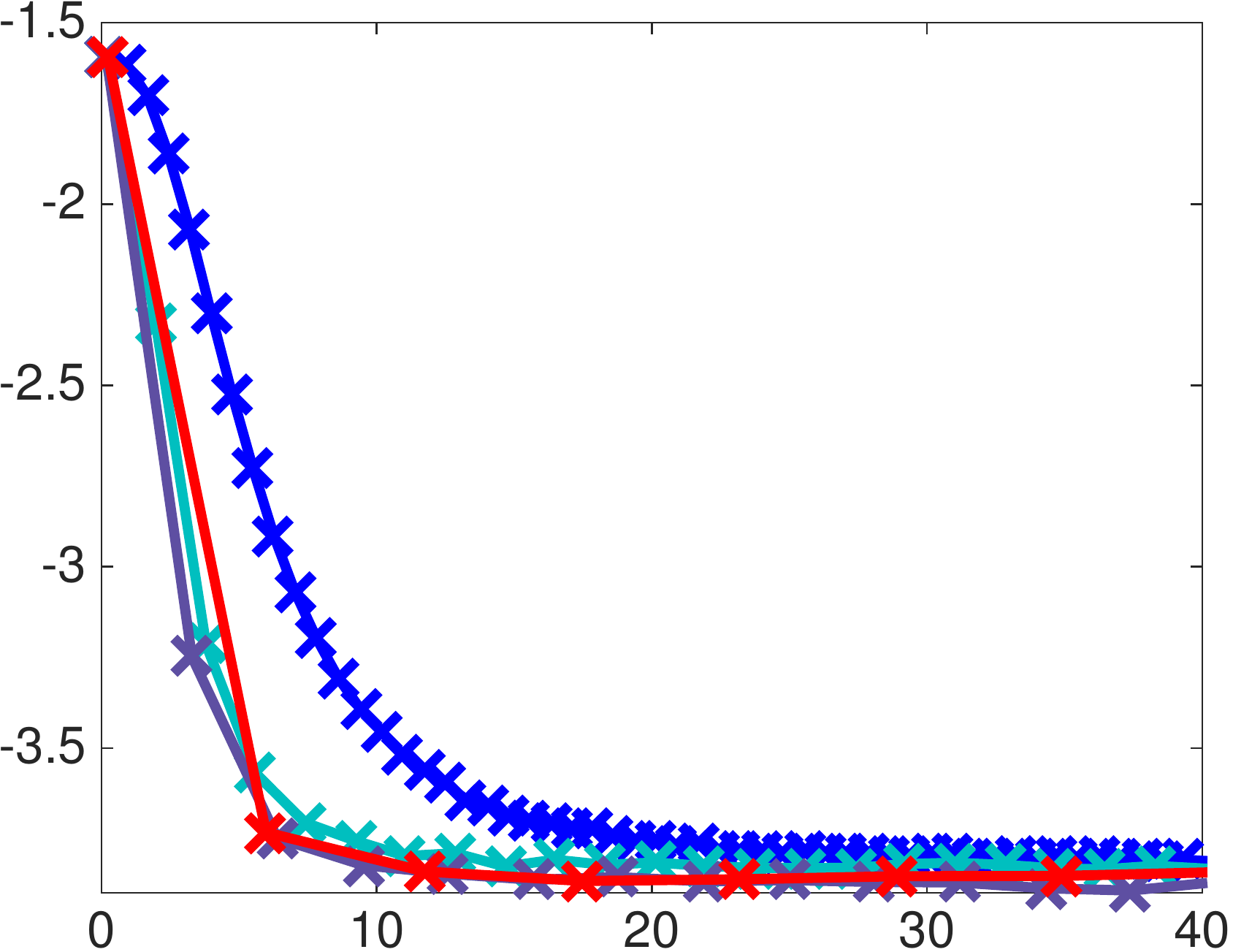} &
\raisebox{2.5em}{\includegraphics[width=0.11\textwidth]{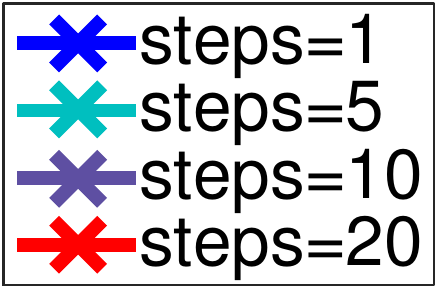}} \\
{\small time (seconds)} & {\small time (seconds)} & {\small time (seconds)} &\\
(a) $\E(x)$ & (b) $\E(x^2)$ & (c)  $\E(\cos(wx+b))$  & \\
\end{tabular}
\vspace{-8bp}\caption{\small
Results when using different numbers of gradient decent steps for solving \eqref{equ:follow2}. 
The setting is the same as that in Figure~\ref{fig:gmm}, but 
we conduct experiments using 1, 5, 10, 20 gradient steps when solving \eqref{equ:follow2}, 
and show their corresponding training time in the $x$-axis, 
and their mean square error for estimating $\E_p h$ (for the ``testing'' distributions) in the $y$-axis. 
%(a)-(c) show the mean square errors for estimating expectation $\E_p(h(x))$ for different $h$ using 
The Langevin samplers we used have $T=10$ layers (Langevin update steps). 
The results are evaluated by drawing 1,000 samples from the trained samplers at different iterations of SVGD. The dimension of the Gaussian Mixtures is $d=50$. 
%Evaluating 
}
%(a)-(c) show the mean square errors estimating expectation $\E_p(h(x))$ for $h(x) = x, x^2$ and $\cos(wx+b)$ where $w\sim \N(0,1)$ and $b\sim \mathrm{Uniform}(0, 2\pi)$ and report the average MSE over 20 random trials.  Langevin ($k$) denotes running $k$ iterations of Langevin dynamics, and Layer ($k$) denotes $T=k$ layer Langevin sampler.}
\label{fig:computation}
\end{figure*}
%\end{comment}

\begin{figure*}[htbp]
\centering
\begin{tabular}{ccc}
\includegraphics[width=0.25\textwidth]{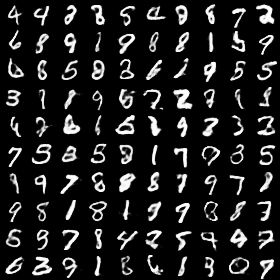} \hspace{20bp} &
\includegraphics[width=0.25\textwidth]{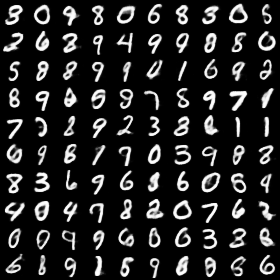} \hspace{20bp} &
\includegraphics[width=0.25\textwidth]{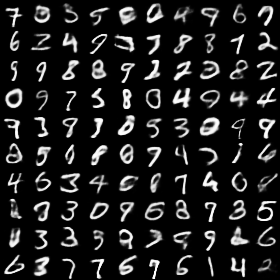} \\
\small VAE-CNN & \small EVAE-CNN & \small ESteinVAE-CNN \\
\end{tabular}
\vspace{-8bp}\caption{\small Images generated by VAE-CNN, EVAE-CNN, and ESteinVAE-CNN}
\label{fig:vae_samples}
\end{figure*}
 
%\begin{comment}
\section{Solving the Projection Step Using Different Numbers Gradient Steps}\label{sec:diffsteps}
Amortized SVGD requires us to solve the projection step 
using either \eqref{equ:follow1} or \eqref{equ:follow2} at each iteration. 
In practice, we approximately solve it using only one step of gradient descent 
starting from the old values of $\eta$ for the sake of computational efficiency. 

In order to study the trade-off of accuracy and computational cost here, 
we plot in Figure~\ref{fig:computation} %, % shows the result 
the results when we solve Eq~\eqref{equ:follow2} using different numbers of gradient descent steps (the result is almost identical when we solve Eq~\eqref{equ:follow1} instead).  
We can see that when using more gradient steps, %of gradient descent, 
although the training time per iteration increases, 
the overall convergence speed may still improve, because it may take less iterations to converge. 
Figure~\ref{fig:computation} seems to suggest that using 5, 10,  20 steps gives better convergence than using a single step, but this may vary in different cases. We suggest to search for the best gradient step if the convergence speed is a primary concern. 
On the other hand, the number of gradient steps seems to have minor influence on the final result at the convergence as shown in Figure~\ref{fig:computation}. 
%we can get slightly better accuracy, but the training time (per iteration) increases significantly.  
%It seems to be reasonable to generally suggest to use a single gradient step as the default choice, 
%although it is possible that using multiple gradient steps can increase the convergent rate. 
%but it may
% that using $1$ gradient step is a good
%Using a proper number steps of gradient descent may vary when applying amortized SVGD to different tasks, which is worth investigation in practice.
%\end{comment}

\section{Images Generated by Different VAEs}
Figure \ref{fig:vae_samples} shows  the images generated by the standard VAE-CNN, the entropy regularized VAE-CNN and ESteinVAE-CNN. We can see that both EVAE-CNN and ESteinVAE-CNN can generate images of good quality.

\end{document}